# Deep Learning-Based Longitudinal Prediction of Childhood Myopia Progression Using Fundus Image Sequences and Baseline Refraction Data

## Authors

**Mengtian Kang[1,10], Yansong Hu[2,10], Shuo Gao[2,10], Ankang Zhou[2], Yuanyuan Liu[3], Hongbei Meng[2], Xuemeng Li[2], Shengbo Wang[2], Xuhang Chen[4], Hubin Zhao[5], Jing Fu[1], Guohua Hu[6], Wei Wang[7], Yanning Dai[8], Arokia Nathan[9], Peter Smielewski[4], Ningli Wang[1], Shiming Li[1]**

## Affiliations

1 Beijing Tongren Hospital, Capital Medical University, 100005, Beijing, China

2 School of Instrumentation and Optoelectronic Engineering, Beihang University, 100191, Beijing, China

3 School of Mechanical Engineering, Beihang University, 100091, Beijing, China

4 Division of Neurosurgery, Department of Clinical Neurosciences, University of Cambridge, CB2 2PY, Cambridge, UK

5 Division of Surgery and Interventional Science, University College London, HA7 4LP, Stanmore, UK

6 Department of Electronic Engineering, The Chinese University of Hong Kong, Shatin, N. T., 999077, Hong Kong S. A. R., China

7 AI Medical Image Division, Thorough Future Inc., 100036, Beijing, China

8 Centers of Excellence, King Abdullah University of Science and Technology, 23955, Thuwal, Saudi Arabia

9 Department of Engineering, University of Cambridge, CB2 1PZ, Cambridge, UK

10 These authors contributed equally: Mengtian Kang, Yansong Hu, Shuo Gao

## Abstract

Childhood myopia constitutes a significant global health concern. It exhibits an escalating prevalence and has the potential to evolve into severe, irreversible conditions that detrimentally impact familial well-being and create substantial economic costs. Contemporary research underscores the importance of precisely predicting myopia progression to enable timely and effective interventions, thereby averting severe visual impairment in children. Such predictions predominantly rely on subjective clinical assessments, which are inherently biased and resource-intensive, thus hindering their widespread application. In this study, we introduce a novel, high-accuracy method for quantitatively predicting the myopic trajectory and myopia risk in children using only fundus images and baseline refraction data. This approach was validated through a six-year longitudinal study of 3,408 children in Henan, utilizing 16,211 fundus images and corresponding refractive data. Our method based on deep learning demonstrated predictive accuracy with an error margin of 0.322D per year and achieved area under the curve (AUC) scores of 0.941 and 0.985 for forecasting the risks of developing myopia and high myopia, respectively. These findings confirm the utility of our model in supporting early intervention strategies and in significantly reducing healthcare costs, particularly by obviating the need for additional metadata and repeated consultations. Furthermore, our method was designed to rely only on fundus images and refractive error data, without the need for meta data or multiple inquiries from doctors, strongly reducing the associated medical costs and facilitating large-scale screening. To serve impoverished areas where people face difficulties in visiting medical institutions, our model can even provide good predictions based on only a single time measurement. Consequently, the proposed method is an important means to reduce medical inequities caused by economic disparities.

**Introduction**

High myopia is a prevalent global health concern, and is anticipated to affect over 1 billion individuals by 2050[1]. This condition and its associated complications, including amblyopia, strabismus, and retinal detachment, significantly increase the risk of severe ophthalmologic conditions such as cataracts and glaucoma, potentially leading to permanent vision loss and representing a primary cause of blindness worldwide. Given the irreversible progression of high myopia, early intervention for at-risk individuals is recognized as the most effective preventive strategy[2,3]. Empirical evidence indicates a strong correlation between the onset of myopia in childhood and the later development of high myopia and related complications during adolescence and adulthood[4]. Consequently, proactive and timely treatment of childhood myopia is crucial to prevent escalation of the disorder into more severe conditions. However, addressing this issue effectively poses significant challenges for the following three reasons.

First, while studies indicate that pharmacologic interventions can reduce myopic progression by over 70% in children aged 6 to 13 years[5], the lack of predictive ability regarding individual myopic trajectories impedes timely and targeted treatment[6]. Second, traditional methods for assessing and predicting myopia trends heavily rely on subjective assessments by specialists, leading to significant variability in predictions and placing a substantial strain on regional medical resources. Third, the prevalence of childhood myopia is alarmingly high and continues to rise, with the World Health Organization reporting global rates of 40% and figures exceeding 90% in some Asian countries such as South Korea[7]. Contemporary research often utilizes machine learning models that incorporate multidimensional data, including lifestyle, physiological, and genetic factors, to forecast myopia risk in children[17,18,19]. However, these methods require extensive data and are currently inadequate for early detection among the vast population in his age group. There is thus a pressing need for scalable techniques capable of accurately predicting myopia progression at early stages to facilitate effective intervention.

Achieving large-scale childhood myopia screening has long been considered a "mission impossible." However, the rapid advancements in imaging technology and artificial intelligence have now made this goal feasible. Fundus imaging technology, which is non-invasive and cost-effective (with specialized equipment priced below $10,000[27], significantly lower than OCT at over $60,000[28] and MRI at over $1,000,000[29]), is ideal for widespread use[8]. Fundus images offer a wealth of deep biomarker data that are crucial for understanding the pathology of myopia. The advent of deep learning has further enabled the extraction and analysis of complex data from these images[7]. Recent systematic reviews and research have documented numerous successful

applications of fundus images in diagnosing serious ocular conditions such as glaucoma, cataracts, and pathological myopia[9,10,11] and in attempts to diagnose general myopia and measure refractive errors using deep learning models[12,13,14]. Preliminary studies are also exploring the use of fundus images alongside baseline physiological data to predict future myopia risk in children[15,16]. Despite these advances, current methods based on fundus images struggle to accurately identify children at high risk of myopia within a short time frame (i.e., within three years) and cannot quantitatively predict the progression of a child's myopia. Additionally, as previously mentioned, traditional methods relying on physiological data necessitate extensive surveys and large datasets, which complicates early screening and the provision of effective interventions.

To address the need for large-scale screening of childhood myopia, this study introduces a novel quantitative prediction method to forecast myopia progression in children. This approach uniquely extracts and integrates in-depth information on myopia progression in individual fundus images and temporal information on myopia progression in fundus image sequences, thereby achieving high-precision predictions of myopia in children. Initially, the proposed method employs a convolutional neural network based on the ResNet34 architecture[20] to extract features from individual fundus images. These features are then integrated at the feature level with corresponding refractive error data through maximal pooling. Subsequently, a recurrent neural network with an LSTM architecture[21] is utilized to analyze temporal dynamics across fundus images, culminating in the prediction of both high myopia risk and quantitative future refractive errors. The dataset for this research was compiled by tracking 3,408 children from Henan Province, China, over six years, resulting in 16,211 fundus images and corresponding refractive data, including 4,863 myopic and 11,348 non-myopic fundus images. The images were annotated with equivalent spherical refractive error data by 23 experts from Beijing Tongren Hospital. Our method achieved an experimental prediction accuracy of 87.0% for myopia risk and 97.9% for high myopia over 5 years, with AUCs of 0.941 and 0.985, respectively, and a mean prediction error of 0.322D for the spherical equivalent refractive error. Detailed results are presented in Table 2 and Figure 2. This method was found to be effective in deriving profound insights into myopia progression from fundus images, and predicting the risk of myopia and high myopia over several years using as little as a single fundus photograph and initial diopter readings. This method can also provide personalized quantitative forecasts for the development trajectory of equivalent spherical refractive errors. The proposed method is poised to become a crucial tool for the extensive screening and management of myopia in children, significantly aiding in reducing medical resource disparities.

## Results

**Study Design and Data Characteristics.** The Anyang Childhood Eye Study (ACES) was a school-based cohort study to determine the annual rate of the incidence of, progression of, and risk factors for myopia among Chinese children in the urban areas of Anyang city, Henan province, and central China. In brief, we recruited 3,408 grade 1 students from 11 randomly selected primary schools from February to May 2012. These students were followed annually for five years. In total, 16,211 macula-centered fundus images with a 45° field of view were utilized to develop and internally validate our deep learning model, as depicted in Figure 1. This model was used to process n (1~5) years of fundus image sequences and corresponding optometric data to predict the future risk of myopia and high myopia, along with refractive errors for m (5~1) subsequent years. Due to incomplete longitudinal data, we partitioned the dataset into training and validation sets in a 5:1 ratio, maintaining consistent feature proportions across both. The internal validation dataset included 56.7% male subjects with baseline myopia and high myopia prevalence of 5.68% and 0.07%, respectively, which increased to 59.27% and 1.90% after five years. We further validated the model's generalizability through an external dataset from Henan, yielding comparable results over a three-year period. Additional demographic and ocular baseline characteristics, as well as five-year follow-up data, are detailed in Supplementary Table 1.

**Method Performance and Model Explainability.** As depicted in Figure 4, our methodology demonstrated robust performance in predicting future myopia progression in children using minimal data. In internal validations, the accuracy for predicting sixth-year myopia risk using 1~5 years image sequences was 72.7%, 76.6%, 84.1%, 87.6%, and 90.5%, respectively. For high myopia, the accuracy was 95.7%, 93.6%, 98.0%, 95.4%, and 96.6%. Details for the other years are presented in Table 2 and Figure 2. The absolute mean errors in predicting myopic equivalent spherical refractive errors over the next 1-5 years using only one year of data were 0.369 D, 0.554 D, 0.704 D, 0.906D, and 1.098 D, respectively; additional data for other durations are also provided in Table 2 and Figure 3. Our method significantly outperformed diagnostics using traditional data, as shown in Table 3. Figure 6 illustrates the model's physiologically relevant focus areas in enhanced fundus images, highlighting key features around the macula and vascular structures near the optic disc. This focus suggests potential paths for future pathological studies on childhood myopia prediction.

**Subgroup Analysis.** We evaluated potential performance biases within different characteristic groups of children, as illustrated in Figure 5. The MAE of quantitative prediction errors for males and females differed by less than 5%, and the predicted ROC curves at different sequence lengths also remained generally consistent. This result suggests that gender has little effect on the performance of our model. For children's initial myopic status, the model's errors in predicting the quantitative future development of myopia for initially myopic children were significantly greater than those for initially nonmyopic children, with a difference of nearly 50%, suggesting that the initial state of myopia has a greater impact on errors in predicting the quantitative development of future myopia. However, when employing relative error measurements, the prediction discrepancies between these groups can also align closely since initially myopic children are themselves more refractive in absolute terms.

**Further Results of Myopia Prediction.** In addition to quantitative regression predictions of myopia progression trajectory, we also provide two different methods for future myopia risk classification predictions. These methods can use the newly trained classifier in the model or directly apply the refractive index prediction results for classification. A comparison of the results under these two different methods can be found in Table 2. Specifically, when the refractive index predictions are used directly for classification, a curve of different predicted myopia versus classification accuracies was obtained by comparing the classification thresholds with the accuracy. As shown in Figure 3, the accuracy values at X=-6 and X=-0.5 on this curve correspond to the predictive ability for high myopia and average myopia risk, respectively. Notably, this curve reflects the distribution in the degree of inconsistency between the model predictions and the true distribution, which offers rich in in-depth information, and the lowest point of the curve, which reflects the most inconsistent categorization thresholds. This curve show the most difficult-to-discriminate and widespread trends in myopia development within this population, which offers insights into regional myopia progression patterns among children.

**Discussion**

In this study, we introduce a novel method to assess the risk of myopia and high myopia and quantitatively predict the developmental trajectory of refractive errors in children. Our approach leverages sequences of fundus images combined with refractive error data to make predictions, achieving remarkable average prediction AUCs of 0.941 and 0.985 for the incidence probability of myopia and high myopia, respectively, along with a mean absolute error (MAE) in SER predictions of approximately 0.3 D per year. Notably, our model requires only fundus images and baseline refractive data, obviating the need for additional patient information such as ocular biological, lifestyle, or environmental parameters. Moreover, our model can generate reliable forecasts from a single clinical encounter. For instance, utilizing one year's data from a single visit can yield AUCs of 0.888 and 0.989 for predicting the risk of myopia and high myopia three years later, with a refractive error prediction MAE of 0.235 D/year. To our knowledge, this is the first methodology that achieves high-precision predictions of myopia and quantitatively charts the progression of SER solely using fundus images and baseline refractive error data. We further elaborate on the technical and practical merits, societal impacts, and other dimensions of our method in subsequent sections.

Our methodology represents a significant advancement in extracting in-depth information from fundus images. While artificial intelligence models based on fundus images have been extensively used to diagnose a variety of ocular diseases such as glaucoma, diabetic retinopathy, and pathological myopia, including both current status assessments and progression predictions[7], myopia remains a challenge due to its higher prevalence and less distinct features compared to other conditions[1]. Unlike diseases such as glaucoma, which is characterized by specific optic nerve and blood vessel structures[22], diabetic retinopathy, marked by microaneurysms and hemorrhages[23], and pathological myopia, which present features such as a tessellated retina[11], normal myopia lacks readily discernible features, complicating the development of accurate prediction models. In response, some researchers have integrated multi-dimensional data, including lifestyle, parental myopia history, and other ocular examinations, into machine learning and decision-making frameworks to enhance myopia risk prediction. For instance, Karla Zadnik et al. successfully forecasted myopia in children aged 7 to 13 using factors such as the baseline spherical equivalent refractive error and parental myopia history, achieving predictive accuracy with an area under the curve between 0.87 and 0.93[24]. Similarly, Haotian Lin et al. utilized longitudinal electronic pathology data from multiple eye centers in China, employing annual myopia progression rates and other metrics to predict high myopia risk over a decade with remarkable precision in external validation datasets and achieving a minimum mean absolute error below 0.3 D, as well as an area under the curve of up to 0.976[19]. However, these approaches often require substantial data and time commitments, leading to significant resource consumption, which poses challenges in medically underserved regions. There have also been efforts to combine fundus images with baseline refractive data to predict high myopia risk. For example, Li Lian Foo et al. employed a hybrid model and achieved an AUC of 0.97 over five years[15]. Nonetheless, these methods generally predict long-term high myopia risks but struggle with providing short-term forecasts or a quantitative trajectory of myopia development.

Our technique advances the use of fundus images through a well-designed preprocessing system (incorporating high-boosting filter, channel transformers, and contrast limited adaptive histogram equalization, as shown in Figure 6 in the Supplementary Information) to accentuate physiological characteristics. This step facilitates feature extraction when using a convolutional neural network, which is further enhanced by a temporal neural network integrating baseline refractive data. Our method successfully captured the underlying deep information related to myopia from the fundus images, as well as temporal information in fundus sequences, enabling the quantitative prediction of general myopia with indistinct symptoms. The results indicate three categories of benefits: data efficiency (requiring minimal data, typically collected in a single visit), predictive versatility (capable of both short-term and long-term forecasts for general and high myopia, while also providing quantitative trajectories of myopia progression), and prediction accuracy (achieving an AUC of 0.941 for myopia and 0.985 for high myopia risk prediction, with an average MAE at 0.322 D/year for the quantitative prediction of spherical equivalent refraction), as illustrated in Figure 4. This methodology not only enhances the feasibility of widespread myopia screening but also reduces the related healthcare burden, offering robust support for the early intervention and management of childhood myopia.

Due to innovative technical advances, our method presents two significant advantages in practical applications. Firstly, this method can be swiftly implemented in regions with limited medical resources, as it requires data from only a single visit, which can save significant time and labor. Additionally, our method exhibits remarkable flexibility. The proposed technique not only captures depth information on myopia progression but also integrates temporal information between fundus sequences to obtain more accurate predictions. Therefore, this method can accommodate varying lengths (ranging from one to five years) of fundus images and equivalent spherical refractive error sequence data. This flexibility allows our approach to fully leverage the existing information from subject children, whether they have prior data or are new subjects, thereby optimizing the use of medical resources. Secondly, through an analysis of fundus photography, our method can identify potential factors and new biomarkers related to the general development of myopia in children. This factor is illustrated by the heat maps generated using the model, as shown in Figure 6. We employed technologies such as Grad-CAM[25] and Guided-Backpropagation[26] to visualize the model heat maps on both the enhanced pre-processed and original fundus images. These heat maps show certain key features considered by the model, such as blood vessel structures connected to the optic disc and peripheral features around the macula. It can be seen that the fundus features related to rapid myopia progression are mainly concentrated in the optic disc, macular area, superior temporal and inferior temporal retinal areas. These results are somewhat consistent with the areas of fundus lesions in high myopia, including maculopathy, dome-shaped macula, myopia-associated glaucoma-like neuropathy, and posterior staphyloma. These characteristics suggest underlying genetic information pertinent to myopia development in children, marking a departure from the biomarkers currently used for myopia diagnosis. Historically, clinicians have struggled to predict myopia progression using fundus images. Our research advances the analysis of myopia characterization in fundus imagery, contributes to ophthalmic pathology research, and supports the discovery of novel biomarkers.

Our deep learning method for predicting childhood myopia significantly enhances the feasibility of large-scale risk screening in primary and rural hospitals with limited medical resources. Unlike traditional screening processes that depend heavily on extensive clinical data from specialized equipment and subjective assessments by experienced ophthalmologists, our approach requires only fundus images and historical refractive data to predict long-term high myopia risk. Thus, this method substantially lowers screening costs and increases operational efficiency. Effective screening enables early intervention for children at high risk of developing high myopia, thereby optimizing resource use in medically underserved areas by reducing unnecessary ophthalmologic consultations for low-risk cases. Additionally, our technique generates personalized, quantitative forecasts of myopia progression (refer to Results Figure 3, where green dashes indicate known data; red, our predictions; and blue, actual outcomes). This predictive capability will not only facilitate timely interventions but also help determine the most appropriate timing, type, and intensity of preventive measures tailored to individual risk profiles, particularly in children identified as high risk. Such targeted interventions are likely to improve the efficacy and reliability of managing potential high myopia, mitigating the escalating global prevalence of myopia among children. Furthermore, our deep learning method to accurately predict childhood myopia developmental trajectories opens avenues for future research, particularly in investigating the correlation between early intervention and future myopic characteristics. This factor represents a pivotal direction for advancing our understanding and management of childhood myopia.

Additionally, our study contributes positively to bridging the urban rural divide and narrowing socioeconomic disparities. Myopia represents a significant global impediment to workforce productivity[30]. In impoverished regions, children often lack timely interventions and treatments for myopia, leading to exacerbated myopia in adulthood. This myopia, profoundly impacts both livelihoods and productivity, thereby perpetuating regional disparities. Relevant data underscore the economic toll of myopia, with East Asia, South Asia, and Southeast Asia experiencing GDP losses of nearing 1.5%. East Asia alone has exceeded $140 billion in productivity losses. Conversely, regions such as Western Europe and Australia, where early myopia receives professional intervention, have experienced GDP losses under 0.1% and productivity losses under $10 billion[31]. Hence, our research offers promise in addressing medical resource disparities and economic imbalances. Long-term myopia or high myopia in childhood is also significantly correlated with psychological disorders, notably contributing to pathological anxiety among teenagers, with a correlation coefficient of $p < 0.05$[32]. Moreover, the progression to high myopia and even pathological myopia markedly increases the incidence of depression and anxiety disorders by over 20%[32]. Additionally, childhood myopia serves as a prominent factor in school bullying, with nearly 48.9% of myopic children fearing ridicule and bullying for wearing glasses, deterring them from wearing corrective eyewear[33]. Given the efficacy of our method in universal screening and developmental predictions, widespread adoption of this method in childhood myopia prevention and intervention holds promise in effectively mitigating school bullying and youth depression.

There are also certain limitations to our study, the most important being the generalization performance of deep learning systems when deployed in clinical settings. Achieving robust generalization necessitates a diverse training dataset. Our study built a prediction model using the same cohort in central China. However, testing the model in other populations will be necessary to evaluate its generalizability, particularly for individuals of East Asian descent. Mitigating these biases would be possible by fine-tuning the model and, using the model with local datasets for retraining and adjustments in future applications. To enhance the model's efficacy, subsequent work should aim to incorporate data encompassing a broader spectrum of regions and ethnicities. Another significant limitation of the present study is the underrepresentation of high myopia in our dataset, which compromises the model's sensitivity to this condition due to its infrequency. Future research should, therefore, prioritize the collection of a more comprehensive dataset to address this imbalance. Additionally, the predominance of children with stable myopia progression in the dataset may introduce a systematic bias against those exhibiting rapid progression. However, this factor is consistent with the general characteristics of the population. Children with severe myopia progression are of significant concern to the model and physicians. In addition, errors for children with higher levels of progress are acceptable when considered from perspective of view of relative errors.

**Methods**

**Study participants.** This school-based cohort study (ACES) recruited 3,408 grade 1 students aged 7.1 $\pm$ 0.4 years (range, 6-9 years) attending 11 randomly selected primary schools from February to May 2012. The selected students were followed annually for five years, resulting in a total of 16,211 fundus images. Each participant underwent a comprehensive dilated pupil optometry test administered by a trained medical professional, with meticulous statistical analysis and documentation conducted thereafter. Fundus images were captured by skilled operators employing a Canon CR-2 non-dilated fundus camera (Canon Inc, Tokyo, Japan). Two fundus photographs of the right eye were taken for each subject, with a 45° field of view, centred around both the optic disc and the macula. For subsequent analysis, this study primarily utilized a fundus photograph of the right eye centred around the macula. All children provided a written informed consent form signed by their parents, and verbal consent was also obtained from each child. This study adhered to the tenets of the Declaration of Helsinki. Ethics committee approval was obtained from the Institutional Review Board of Beijing Tongren Hospital, Capital Medical University.

**Definitions of myopia.** Spherical equivalent refraction (SER, sometimes abbreviated to SE) was defined as the sum of spherical power and half of the cylinder. In this study, myopia was characterized and classified based on the SER to reflect the severity and progression of the disorder. Myopia was defined as a SER of -0.5 D or less, individuals with an SER below -6.0 diopters (D) were categorized as having high myopia, those with an SER between -6.0 D and -3.0 D were classified as having moderate myopia, individuals with an SER between -3.0 D and -0.5 D were categorized as having low myopia, those with an SE between -0.5 D and +3.0 D were classified as emmetropic or having low hyperopia, and those with an SER above +3.0 D were considered hyperopic[48]. The annual progression of myopia was defined as the change in cycloplegic SER between the measurements acquired in the previous year and the measurements taken during the annual follow-up period. Rapid myopia progression was defined as an increase in the myopic spherical equivalent of more than 0.75 D/year among myopic children, while those with an average annual SER progression of less than 0.50 D were labelled as having non-progressive myopia.

**Image preprocessing and enhancement technology.** To ensure the quality of the training data, we developed an image preprocessing and enhancement system, as illustrated in Figure 6 of the Supplementary Information. This system was tailored for the model's requirements. Initially, considering the model's memory constraints, and the clarity and informational content of the images, we standardized the input size to 512 * 512 * 3 through cropping and scaling. Additionally, the original fundus images sometimes contained various defects, such as exposure abnormalities, colour irregularities, and incomplete optic discs, as depicted in Figure 6 of the Supplementary Information. Such defects can adversely impact subsequent training outcomes. Hence, we manually devised three criteria for image filtering: the bright area proportion, representing pixels with grey values greater than the average plus three times the standard deviation (not exceeding 255); the dark area proportion, denoting pixels with grey values lower than the average minus the standard deviation (not less than 5); and the red-blue difference, indicating the disparity between the grey sum of pixels in the red channel and the grey sum in the blue channel. The formula for computing these filtering conditions is presented in Figure 6 of the Supplementary Information. To enhance physiological features such as the optic disc and blood vessels in fundus images while mitigating brightness and contrast disparities among different images (as illustrated in Figure 6 of the Supplementary Information), we also implemented a novel feature enhancement algorithm. This algorithm combines CLAHE (Contrast Constrained Local Histogram Equalization) and High

Pass Filtering techniques. Specifically, this algorithm involves subtracting a Gaussian filter from the original image to derive a high-frequency feature mask, multiplying the mask by a factor of 4, adding the mask to the original image, and subsequently normalizing the grey values and subtracting them from the original image to obtain the final feature-enhanced image. Lastly, the images underwent random horizontal or vertical flipping and normalization before being fed into the model, enhancing the model's robustness.

**Segmentation and distribution characteristics of datasets.** To guarantee the reproducibility of our findings, we employed a consistent random seed to partition the dataset randomly into a training set and a test set, while preserving the underlying distributional properties of the data. For temporal prediction tasks across varying years, we systematically segregated sequences of differing lengths into training and validation sets in a 5:1 ratio, ensuring that key data attributes, such as the prevalence of myopia, remained constant.

**Development of the deep-learning algorithms.** We developed a multi-year myopia prediction network (MMPN) by implementing a model integration strategy. As shown in Figure 5 of the Supplementary Information, in the MMPN, model 1 belonged to the encoder component, through which features are extracted from the input image sequence using the ResNet34 structure based on the CNN architecture and retained in the final convolutional layer. These feature maps are then converted into feature vectors using global average pooling and spliced with the refractive index sequence feature values as the final output of the encoding part, which also serves as the input for the decoding component. Model 2 belongs to the decoding component, which utilizes the LSTM structure in the RNN architecture to process temporal information and perform predictions, outputting the future refractive index sequence. For myopia or high myopia risk predictions, the output must also pass through the fully connected layer classifier, which outputs the risk classification result. The model was trained using Python 1.12.0 as a whole and accelerated using an A5000-24G graphics card. The batch size was set to 8 for the training set and 2 for the validation and test sets. The training process is divided into three phases with learning rates of 1e-3, 1e-4, and 1e-5, for which the number of training rounds was 40, 20, and 10, respectively. The CNN skeleton in model 1 used pre-trained ResNet34 parameters from ImageNet data to initialize the network parameters. The optimization algorithm used Adam's algorithm, the weight-decay parameter was set to 0 in the first period and 1e-4 in the last two periods, and the other parameters were set using the defaults.

**Evaluating the algorithms.** To evaluate the overall performance of the model, we used two types of basic metrics: 1) the accuracy, specificity, and sensitivity, alongside the ROC curves and AUC areas for predicting the risk of future myopia and high myopia in children and 2) the MAE error for quantitatively predicting refractive error progression[49]. The results of the evaluation are shown in Table 2.

**Sub-group analysis.** To assess the effects of factors such as age, gender, and myopia on model performance, we conducted a subgroup analysis. We divided the data into male and female groups, as well as myopic and non-myopic groups, for model validation. The results of the subgroup analysis are shown in Figure 5.

**Model explanation.** When diagnosing general myopia in children and predicting the degree of development, we used visual interpretation tools to understand which regions in the fundus image had the greatest impact on our deep learning model, as shown in Figure 6. We used techniques such as gradient CAM and guided backpropagation to draw weight heatmaps for the model. These maps were used to predict future high myopia risk and quantitatively predict the refractive index. These two technologies offer different methods to attribute some of the network's contributions to the regions in the image. This flexibility enables our model to provide better support for clinical decision-making among doctors, while also contributing to the development of related medical and pathological research.

**Data availability.** Primary fundus images supporting the results of this study, labelling data, and more specific optometric results and medical record data are available upon request from the corresponding author.

**Code availability.** Our deep learning network model was developed using the standard model library and script in Python 1.12.0. The code for fundus image preprocessing technology and deep learning network model construction and training verification was published on GitHub: https://github.com/19376357/Myopia-prediction-model.

# References

1. Holden, B. A. et al. Global Prevalence of Myopia and High Myopia and Temporal Trends from 2000 through 2050. Ophthalmology 123, 1036–1042 (2016).

2. Morgan, I. G., Ohno-Matsui, K. & Saw, S.-M. Myopia. Lancet Lond. Engl. 379, 1739–1748 (2012).

3. Tan, M.-S. et al. Efficacy and adverse effects of ginkgo biloba for cognitive impairment and dementia: a systematic review and meta-analysis. J. Alzheimers Dis. 43, 589–603 (2015).

4. Lanca, C. et al. Rapid myopic progression in childhood is associated with teenage high myopia. Invest. Ophthalmol. Vis. Sci. 62, 17–17 (2021).

5. Ang, M. et al. myopia control strategies recommendations from the 2018 WHO/IAPB/BHVI meeting on myopia. Br. J. Ophthalmol. 104, 1482–1487 (2020).

6. Dhiman, R., Rakheja, V., Gupta, V. & Saxena, R. Current concepts in the management of childhood myopia. Indian J. Ophthalmol. 70, 2800 (2022).

7. Li, T. et al. Applications of deep learning in fundus images: A review. Med. Image Anal. 69, 101971 (2021).

8. Prashar, J. & Tay, N. Performance of artificial intelligence for the detection of pathological myopia from colour fundus images: a systematic review and meta-analysis. Eye 38, 303–314 (2024).

9. Zedan, M. J. et al. Automated glaucoma screening and diagnosis based on retinal fundus images using deep learning approaches: A comprehensive review. Diagnostics 13, 2180 (2023).

10. Junayed, M. S., Islam, M. B., Sadeghzadeh, A. & Rahman, S. CataractNet: An automated cataract detection system using deep learning for fundus images. IEEE Access 9, 128799–128808 (2021).

11. Lu, L. et al. Development of deep learning-based detecting systems for pathologic myopia using retinal fundus images. Commun. Biol. 4, 1225 (2021).

12. Li, J. et al. Automated detection of myopic maculopathy from color fundus photographs using deep convolutional neural networks. Eye Vis. 9, 13 (2022).

13. Zou, H. et al. Identification of ocular refraction based on deep learning algorithm as a novel retinoscopy method. Biomed. Eng. OnLine 21, 87 (2022).

14. Xu, D. et al. Deep learning for predicting refractive error from multiple photorefraction images. Biomed. Eng. OnLine 21, 55 (2022).

15. Foo, L. L. et al. Deep learning system to predict the 5-year risk of high myopia using fundus imaging in children. NPJ Digit. Med. 6, 10 (2023).

16. Wan, C., Li, H., Cao, G.-F., Jiang, Q. & Yang, W.-H. An artificial intelligent risk classification method of high myopia based on fundus images. J. Clin. Med. 10, 4488 (2021).

17.    Yang, X. et al. Prediction of myopia in adolescents through machine learning methods. Int. J. Environ. Res. Public. Health 17, 463 (2020).

18.    Han, X., Liu, C., Chen, Y. & He, M. Myopia prediction: a systematic review. Eye 36, 921–929 (2022).

19.    Lin, H. et al. Prediction of myopia development among Chinese school-aged children using refraction data from electronic medical records: a retrospective, multicentre machine learning study. PLoS Med. 15, e1002674 (2018).

20.    He, K., Zhang, X., Ren, S. & Sun, J. Deep residual learning for image recognition. in Proceedings of the IEEE conference on computer vision and pattern recognition 770–778 (2016).

21.    Graves, A. Long Short-Term Memory. in Supervised Sequence Labelling with Recurrent Neural Networks vol. 385 37–45 (Springer Berlin Heidelberg, Berlin, Heidelberg, 2012).

22.    Norouzifard, M. et al. Identification of clinically relevant glaucoma biomarkers on fundus images using deep learning. Invest. Ophthalmol. Vis. Sci. 60, PB090–PB090 (2019).

23.    Cunha-Vaz, J., Ribeiro, L. & Lobo, C. Phenotypes and biomarkers of diabetic retinopathy. Prog. Retin. Eye Res. 41, 90–111 (2014).

24.    Zadnik, K. et al. Prediction of juvenile-onset myopia. JAMA Ophthalmol. 133, 683–689 (2015).

25.    Selvaraju, R. R. et al. Grad-cam: Visual explanations from deep networks via gradient-based localization. in Proceedings of the IEEE international conference on computer vision 618–626 (2017).

26.    Gu, J., Yang, Y. & Tresp, V. Understanding Individual Decisions of CNNs via Contrastive Backpropagation. in Computer Vision – ACCV 2018 (eds. Jawahar, C. V., Li, H., Mori, G. & Schindler, K.) vol. 11363 119–134 (Springer International Publishing, Cham, 2019).

27.    Das, S. et al. Very low price smartphone fundus camera innovation with basic amenities– a novel approach toward undergraduate teaching. Med. J. Dr DY Patil Univ. 14, 592–595 (2021).

28.    Song, G. et al. First clinical application of low-cost OCT. Transl. Vis. Sci. Technol. 8, 61–61 (2019).

29.    Bradley, W. Comparing costs and efficacy of MRI. Am. J. Roentgenol. 146, 1307–1310 (1986).

30.    Sankaridurg, P. et al. IMI impact of myopia. Invest. Ophthalmol. Vis. Sci. 62, 2–2 (2021).

31.    Naidoo, K. S. et al. Potential lost productivity resulting from the global burden of myopia: systematic review, meta-analysis, and modeling. Ophthalmology 126, 338–346 (2019).

32.    Łazarczyk, J. B. et al. The differences in level of trait anxiety among girls and boys aged 13–17 years with myopia and emmetropia. BMC Ophthalmol. 16, 201 (2016).

33.    Morjaria, P., Evans, J. & Gilbert, C. Predictors of spectacle wear and reasons for nonwear in students randomized to ready-made or custom-made spectacles: results of secondary objectives from a randomized noninferiority trial. JAMA Ophthalmol. 137, 408–414 (2019).

34. Zadnik, K. et al. Ocular predictors of the onset of juvenile myopia. Invest. Ophthalmol. Vis. Sci. 40, 1936–1943 (1999).

35. Jones, L. A. et al. Parental history of myopia, sports and outdoor activities, and future myopia. Invest. Ophthalmol. Vis. Sci. 48, 3524–3532 (2007).

36. Zhang, M. et al. Validating the accuracy of a model to predict the onset of myopia in children. Invest. Ophthalmol. Vis. Sci. 52, 5836–5841 (2011).

37. French, A. N., Morgan, I. G., Mitchell, P. & Rose, K. A. Risk factors for incident myopia in Australian schoolchildren: the Sydney adolescent vascular and eye study. Ophthalmology 120, 2100–2108 (2013).

38. Chua, S. Y. L. et al. Age of onset of myopia predicts risk of high myopia in later childhood in myopic Singapore children. Ophthalmic Physiol. Opt. 36, 388–394 (2016).

39. Ma, Y. et al. Cohort study with 4‐year follow‐up of myopia and refractive parameters in primary schoolchildren in Baoshan District, Shanghai. Clin. Experiment. Ophthalmol. 46, 861–872 (2018).

40. Tideman, J. W. L., Polling, J. R., Jaddoe, V. W., Vingerling, J. R. & Klaver, C. C. Environmental risk factors can reduce axial length elongation and myopia incidence in 6- to 9-year-old children. Ophthalmology 126, 127–136 (2019).

41. Chen, Y. et al. Contribution of genome-wide significant single nucleotide polymorphisms in myopia prediction: findings from a 10-year cohort of Chinese twin children. Ophthalmology 126, 1607–1614 (2019).

42. Wong, Y. L. et al. Prediction of myopia onset with refractive error measured using non-cycloplegic subjective refraction: the WEPrOM Study. BMJ Open Ophthalmol. 6, e000628 (2021).

43. Shen, Y. et al. Big-data and artificial-intelligence-assisted vault prediction and EVO-ICL size selection for myopia correction. Br. J. Ophthalmol. 107, 201–206 (2023).

44. Huang, J., Ma, W., Li, R., Zhao, N. & Zhou, T. Myopia prediction for children and adolescents via time-aware deep learning. Sci. Rep. 13, 5430 (2023).

45. Guo, C. et al. Development and validation of a novel nomogram for predicting the occurrence of myopia in schoolchildren: A prospective cohort study. Am. J. Ophthalmol. 242, 96–106 (2022).

46. Manoharan, M. K. et al. Myopia progression risk assessment score (MPRAS): A promising new tool for risk stratification. Sci. Rep. 13, 8858 (2023).

47. Li, J. et al. Accurate prediction of myopic progression and high myopia by machine learning. Precis. Clin. Med. 7, pbae005 (2024).

48. Wang, J. et al. How to Conduct School Myopia Screening: Comparison Among Myopia Screening Tests and Determination of Associated Cutoffs. Asia-Pac. J. Ophthalmol. 11, 12 (2022).

49.     Zhao, J. et al. Development and validation of predictive models for myopia onset and progression using extensive 15-year refractive data in children and adolescents. J. Transl. Med. 22, 289 (2024).

# 1 Tables

| | | Model | Num | Age (d) | Sex (M%) | DS (D) | DC (D) | Axis (°) | CT (μm) | AL (mm) | SER(D) | Mild Myopia | Moderate and High Myopia |
|---|---|---|---|---|---|---|---|---|---|---|---|---|---|
| Baseline | | | 2279 | 2614 | 42.78% | 1.212 | -0.506 | 99.82 | 539.96 | 22.71 | 0.964 | 5.70% | 0.48% |
| After 5 years | | | 1940 | 4436 | 44.65% | -1.07 | -0.539 | 103.84 | 546.65 | 24.17 | -1.356 | 40.00% | 21.34% |
| The situation of predictive models for sequences of different lengths | Training Set | 1p1 | 7748 | 3309 | 44.14% | 0.36 | -0.465 | 90.59 | 543.47 | 23.31 | 0.129 | 21.12% | 5.01% |
| | | 1p2 | 5526 | 3124 | 44.79% | 0.633 | -0.454 | 90.48 | 542.58 | 23.14 | 0.402 | 16.61% | 2.70% |
| | | 1p3 | 3532 | 2922 | 44.48% | 0.921 | -0.459 | 90.67 | 542.22 | 23.02 | 0.699 | 11.41% | 0.88% |
| | | 1p4 | 2146 | 2766 | 44.22% | 1.09 | -0.471 | 90.56 | 541.24 | 22.87 | 0.86 | 7.36% | 0.38% |
| | | 1p5 | 989 | 2619 | 43.88% | 1.271 | -0.498 | 93.99 | 538.93 | 22.69 | 1.023 | 4.55% | 0.30% |
| | | 2p1 | 4788 | 3526 | 43.98% | 0.13 | -0.458 | 87.4 | 544.55 | 23.5 | -0.095 | 25.23% | 6.42% |
| | | 2p2 | 3016 | 3318 | 43.40% | 0.425 | -0.441 | 87.88 | 544.17 | 23.35 | 0.214 | 21.32% | 3.35% |
| | | 2p3 | 1746 | 3143 | 44.73% | 0.679 | -0.444 | 86.89 | 543.53 | 23.24 | 0.477 | 16.61% | 1.26% |
| | | 2p4 | 701 | 2956 | 45.22% | 0.931 | -0.44 | 83.97 | 543.55 | 23.22 | 0.717 | 11.55% | 0.14% |
| | | 3p1 | 2844 | 3694 | 44.02% | -0.066 | -0.46 | 89.6 | 545.11 | 23.61 | -0.294 | 29.15% | 7.81% |
| | | 3p2 | 1661 | 3514 | 44.61% | 0.182 | -0.447 | 89.98 | 544.82 | 23.45 | -0.04 | 27.27% | 4.88% |
| | | 3p3 | 659 | 3327 | 44.46% | 0.487 | -0.443 | 90.06 | 543.48 | 23.31 | 0.288 | 20.94% | 2.73% |
| | | 4p1 | 1571 | 3886 | 43.79% | -0.337 | -0.457 | 89.89 | 545.74 | 23.78 | -0.557 | 32.34% | 10.37% |
| | | 4p2 | 633 | 3703 | 43.76% | -0.041 | -0.436 | 88.3 | 543.53 | 23.59 | -0.265 | 28.91% | 7.74% |
| | | 5p1 | 597 | 4069 | 44.39% | -0.618 | -0.49 | 91.44 | 544.48 | 23.9 | -0.832 | 35.68% | 14.40% |
| | Validation Set | 1p1 | 1549 | 3320 | 41.79% | 0.356 | -0.478 | 89.99 | 543.5 | 23.29 | 0.107 | 20.21% | 5.42% |
| | | 1p2 | 1105 | 3122 | 40.54% | 0.647 | -0.466 | 90.42 | 543.69 | 23.2 | 0.411 | 17.10% | 2.44% |
| | | 1p3 | 706 | 2914 | 40.23% | 0.854 | -0.463 | 90.92 | 540.37 | 22.98 | 0.62 | 13.03% | 0.71% |
| | | 1p4 | 429 | 2763 | 43.82% | 1.116 | -0.449 | 92.99 | 539.43 | 22.95 | 0.888 | 8.63% | 0.23% |
| | | 1p5 | 198 | 2625 | 47.98% | 1.215 | -0.454 | 107.96 | 538.42 | 22.69 | 0.99 | 6.06% | 0 |
| | | 2p1 | 957 | 3526 | 45.04% | 0.113 | -0.45 | 92.39 | 545.5 | 23.45 | -0.124 | 26.12% | 6.48% |
| | | 2p2 | 603 | 3322 | 46.43% | 0.485 | -0.431 | 89.34 | 544.77 | 23.28 | 0.277 | 19.90% | 2.49% |
| | | 2p3 | 350 | 3132 | 40.86% | 0.814 | -0.435 | 86.52 | 546.84 | 23.2 | 0.575 | 14.29% | 1.43% |
| | | 2p4 | 140 | 2977 | 39.29% | 0.96 | -0.455 | 75.61 | 543.58 | 22.99 | 0.732 | 12.14% | 0.71% |
| | | 3p1 | 569 | 3674 | 44.46% | -0.057 | -0.433 | 93.78 | 542.77 | 23.57 | -0.254 | 29.35% | 7.38% |

| | | | | | | | | | | | |
|---|---|---|---|---|---|---|---|---|---|---|---|
| 3p2 | 332 | 3514 | 42.47% | 0.294 | -0.391 | 86.43 | 541.86 | 23.43 | 0.122 | 20.78% | 3.92% |
| 3p3 | 132 | 3309 | 44.70% | 0.51 | -0.49 | 83.4 | 542.21 | 23.24 | 0.264 | 26.52% | 1.52% |
| 4p1 | 314 | 3889 | 47.45% | -0.318 | -0.492 | 94.23 | 542.63 | 23.71 | -0.539 | 31.85% | 11.78% |
| 4p2 | 126 | 3698 | 48.41% | -0.293 | -0.414 | 90.49 | 548.77 | 23.7 | -0.495 | 37.30% | 7.14% |
| 5p1 | 119 | 4090 | 47.06% | -0.373 | -0.486 | 84 | 42.1 | 23.89 | -0.534 | 31.09% | 10.92% |

*Num, number of pictures; M, male; DS, diopter of spherical power; DC, diopter of cylindrical power; CT, corneal thickness; AL, axial length; SER, spherical equivalent refraction; NpM, Model predicting future year m with n-year series. All data are arithmetic averages.

**Table 1 Baseline characteristics and Data Summary.** For various input and output sequence lengths, the model was segmented and trained on the dataset separately, and the corresponding baseline feature distributions are provided.

| Model | Future SER quantitative prediction | | Myopia Risk Prediction | | | | | | | High Myopia Risk Prediction | | | | | | |
|---|---|---|---|---|---|---|---|---|---|---|---|---|---|---|---|---|
| | | | Trained Classifier | | | | Threshold Classifier | | | Trained Classifier | | | | Threshold Classifier | | |
| | MAE/D | R2 | Accuracy | Sensitivity | Specificity | AUC | Accuracy | Sensitivity | Specificity | Accuracy | Sensitivity | Specificity | AUC | Accuracy | Sensitivity | Specificity |
| 1p1 | 0.369 | 0.904 | 91.30% | 91.66% | 90.61% | 0.973 | 92.42% | 91.84% | 93.49% | 99.19% | 99.18% | 100.0% | 0.999 | 98.43% | 99.52% | 89.01% |
| 1p2 | 0.554 | 0.971 | 86.64% | 86.93% | 86.27% | 0.933 | 88.27% | 87.28% | 89.63% | 98.64% | 98.63% | 100.0% | 0.991 | 99.52% | 99.75% | 90.25% |
| 1p3 | 0.704 | 0.709 | 78.71% | 72.21% | 84.26% | 0.888 | 82.29% | 81.73% | 83.05% | 98.43% | 98.55% | 87.5% | 0.989 | 98.00% | 99.15% | 65.05% |
| 1p4 | 0.906 | 0.66 | 75.79% | 72.29% | 79.10% | 0.858 | 79.79% | 74.02% | 84.13% | 97.26% | 97.42% | 88.9% | 0.928 | 98.21% | 99.02% | 21.78% |
| 1p5 | 1.098 | 0.549 | 72.73% | 68.83% | 75.45% | 0.782 | 76.47% | 68.66% | 80.83% | 95.72% | 95.70% | 100.0% | 0.911 | 99.12% | 100.0% | 0.00% |
| 2p1 | 0.352 | 0.93 | 90.90% | 87.64% | 95.60% | 0.973 | 91.40% | 90.86% | 92.11% | 98.30% | 98.28% | 100.0% | 0.997 | 99.74% | 100.0% | 83.33% |
| 2p2 | 0.551 | 0.864 | 87.08% | 84.90% | 89.93% | 0.944 | 87.24% | 86.50% | 88.05% | 97.74% | 97.88% | 83.3% | 0.981 | 99.17% | 99.56% | 43.67% |
| 2p3 | 0.773 | 0.687 | 83.75% | 79.43% | 87.11% | 0.920 | 81.50% | 76.80% | 85.39% | 96.75% | 96.73% | 100.0% | 0.967 | 97.86% | 100.0% | 60.05% |
| 2p4 | 0.866 | 0.676 | 76.60% | 66.27% | 91.38% | 0.885 | 79.43% | 84.62% | 77.45% | 93.62% | 95.52% | 57.1% | 0.971 | 94.32% | 95.53% | 0.00% |
| 3p1 | 0.423 | 0.913 | 90.94% | 89.00% | 93.01% | 0.974 | 88.45% | 83.06% | 94.66% | 98.22% | 98.22% | 100.0% | 0.996 | 98.05% | 98.65% | 67.50% |
| 3p2 | 0.542 | 0.879 | 92.37% | 95.89% | 90.28% | 0.976 | 87.79% | 85.00% | 90.14% | 96.44% | 96.40% | 100.0% | 0.991 | 97.88% | 98.01% | 75.00% |
| 3p3 | 0.629 | 0.84 | 84.11% | 82.46% | 85.11% | 0.932 | 84.11% | 84.78% | 83.81% | 98.01% | 97.93% | 100.0% | 0.932 | 98.01% | 98.77% | 40.00% |
| 4p1 | 0.187 | 0.969 | 91.17% | 87.50% | 93.53% | 0.962 | 84.82% | 77.41% | 93.17% | 95.32% | 95.31% | 100.0% | 0.999 | 99.42% | 99.42% | 100.0% |
| 4p2 | 0.559 | 0.877 | 87.60% | 88.64% | 87.06% | 0.939 | 89.15% | 84.31% | 92.31% | 95.35% | 95.31% | 100.0% | 0.977 | 96.87% | 98.89% | 40.00% |
| 5p1 | 0.349 | 0.943 | 90.52% | 97.50% | 86.84% | 0.989 | 91.38% | 85.11% | 95.65% | 96.55% | 96.52% | 100.0% | 0.996 | 98.35% | 99.04% | 67.78% |
| Avg/y | 0.322 | 0.86 | 86.98% | 85.19% | 88.90% | 0.941 | 87.41% | 85.28% | 89.62% | 97.89% | 97.94% | 96.18% | 0.985 | 98.63% | 99.35% | 69.98% |

**Table 2 Model Performance.** NpM refers to the prediction of m years in the future using a sequence of n years, and the table shows the performance of our classification task for future myopia risk prediction and the performance of the regression task in quantitatively predicting future refractive error development. The trained classifier refers to the use of deep learning techniques to classify the final features of our model, while the threshold classifier refers to the use of our quantitative prediction of myopia progression to directly classify future myopia risk using thresholds. The final average was obtained by considering the amount of data from the different models and the weighted average of the forecast years.

|  | MAE/D | Accuracy | Specificity | Sensitivity | AUC |
|---|---|---|---|---|---|
| Our method | 0.32 | 87.0% | 85.2% | 88.9% | 0.94 |
| Logistic Regression | | | | | |
| Baseline SER | 0.85 | 74.9% | 83.2% | 63.0% | 0.82 |
| Anterior chambers depth | 1.25 | 60.0% | 73.1% | 73.2% | 0.64 |
| Axial length/ corneal radius of curvature | 0.77 | 66.9% | 74.5% | 56.3% | 0.72 |
| Six variables[*] | 0.63 | 79.1% | 82.5% | 74.8% | 0.84 |
| Random Forest（Six variables[*]） | 0.75 | 80.0% | 81.1% | 81.2% | 0.84 |

**Table 3 Comparison of methods.** A comparison of our method with the baseline methods. Our method offers better accuracy, and a better AUC compared than all other methods.

[*]Uncorrected distance visual acuity, baseline SER, axial length, corneal radius of curvature, gender, parental myopia.

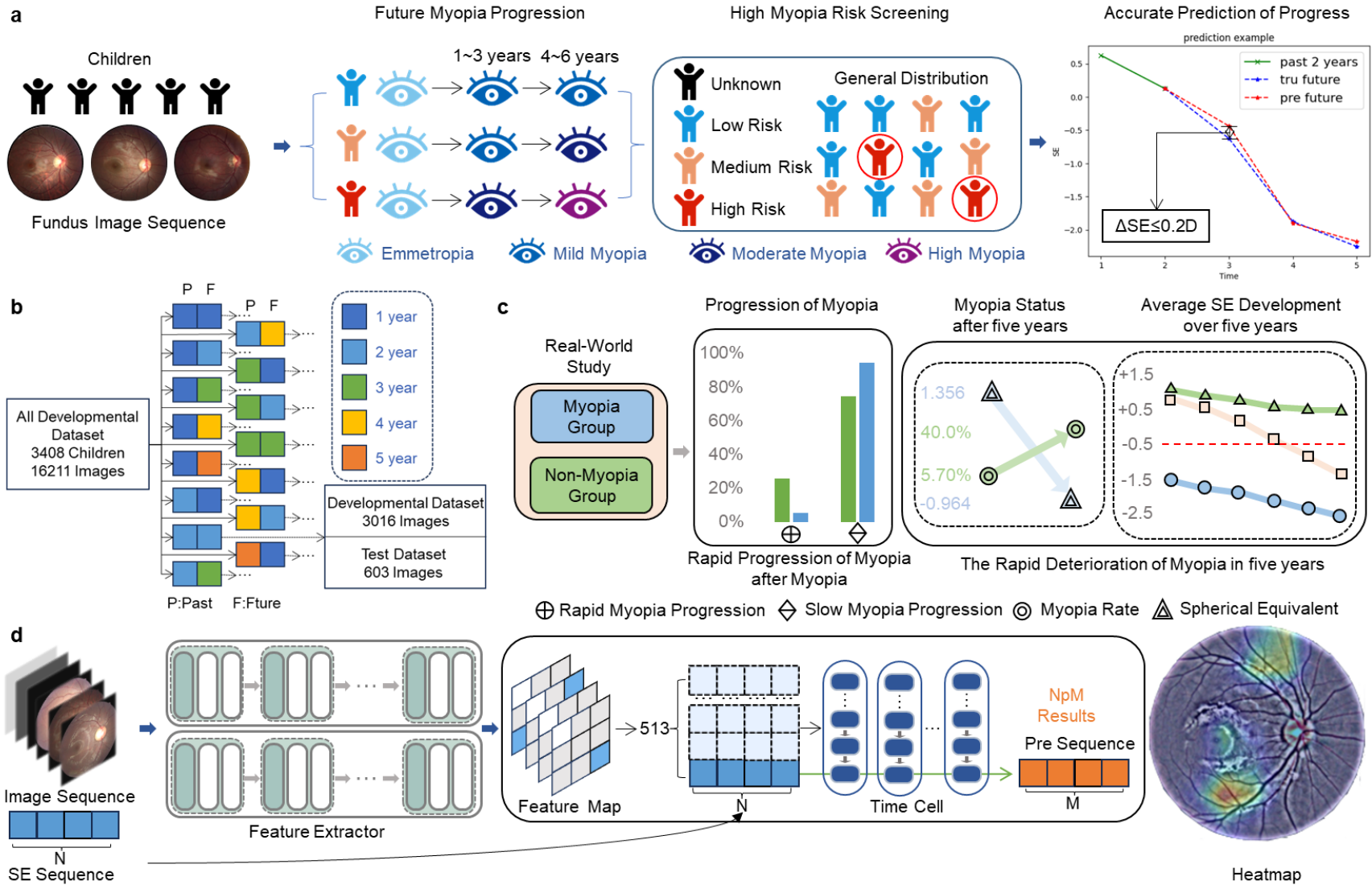

**Fig.1 Study design. a**, System schematic. The system receives children's fundus image sequences and can screen out the high-risk children who have a high probability of future myopia and fast myopia progression. The system can also quantitatively predict the myopia progression of children in the next few years. **b**, Data segmentation schematic. Six consecutive years of data for the children are classified into 15 categories according to the known and predicted years, for example, using the 2 years of fundus sequences to predict the myopia progression of the following 4 years, and each category is classified into 15 categories according to the known and predicted years. For example, the 2-year fundus series data were used to predict myopia development in the following 4 years, and each category was further divided into support data and test data. **c**, The actual examination of the children's population characteristics. Comparing myopia development between originally myopic and originally non-myopic children, and the data study of myopia development in children together showed that the children's myopia development was very rapid over five years. **d**, Model schematic. The model uses n years of fundus image sequences and SE sequences for feature extraction and time series modelling to predict myopia over the next m years, and is used to create the model heat map. See Figures 5 and 6 of the Supplementary Material for model details and preprocessing system details.

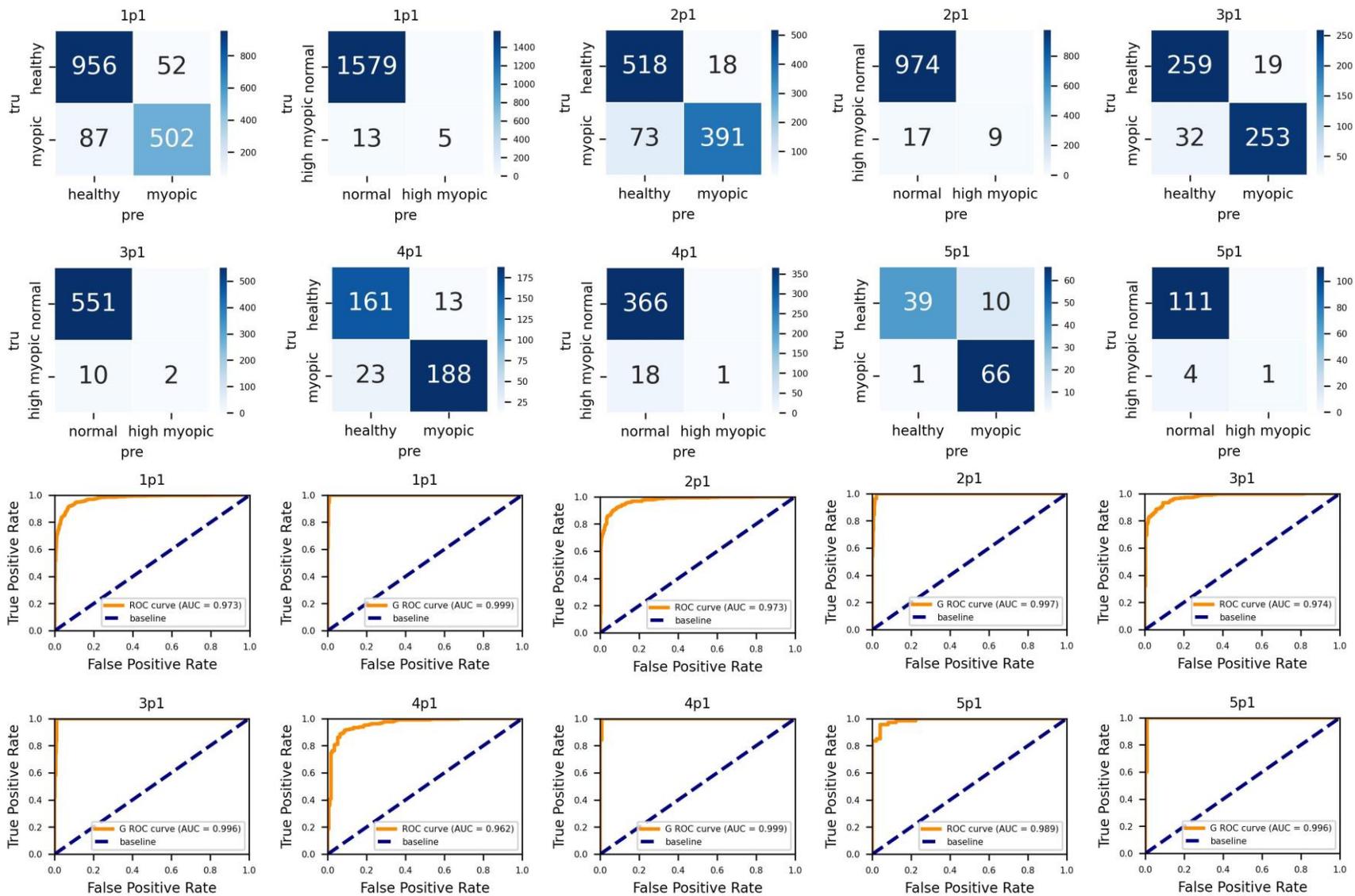

**Fig.2 Results of myopia and high myopia risk prediction.** We used confusion matrices, ROC curves, and other methods to characterize the results and performance of the model. This figure provides an example of a prediction for the following year. See the 59  supporting information for other results. It can be seen that the AUC for predicting the risk of both myopia and high myopia in the
60  coming year can reach more than 0.96. (The npm at the top of the picture indicates that the n-year sequence is used to predict the
61  future m-year sequence, and the same is true for the subsequent ones as well as in SI)
62
63
64
65
66
67
68
69
70

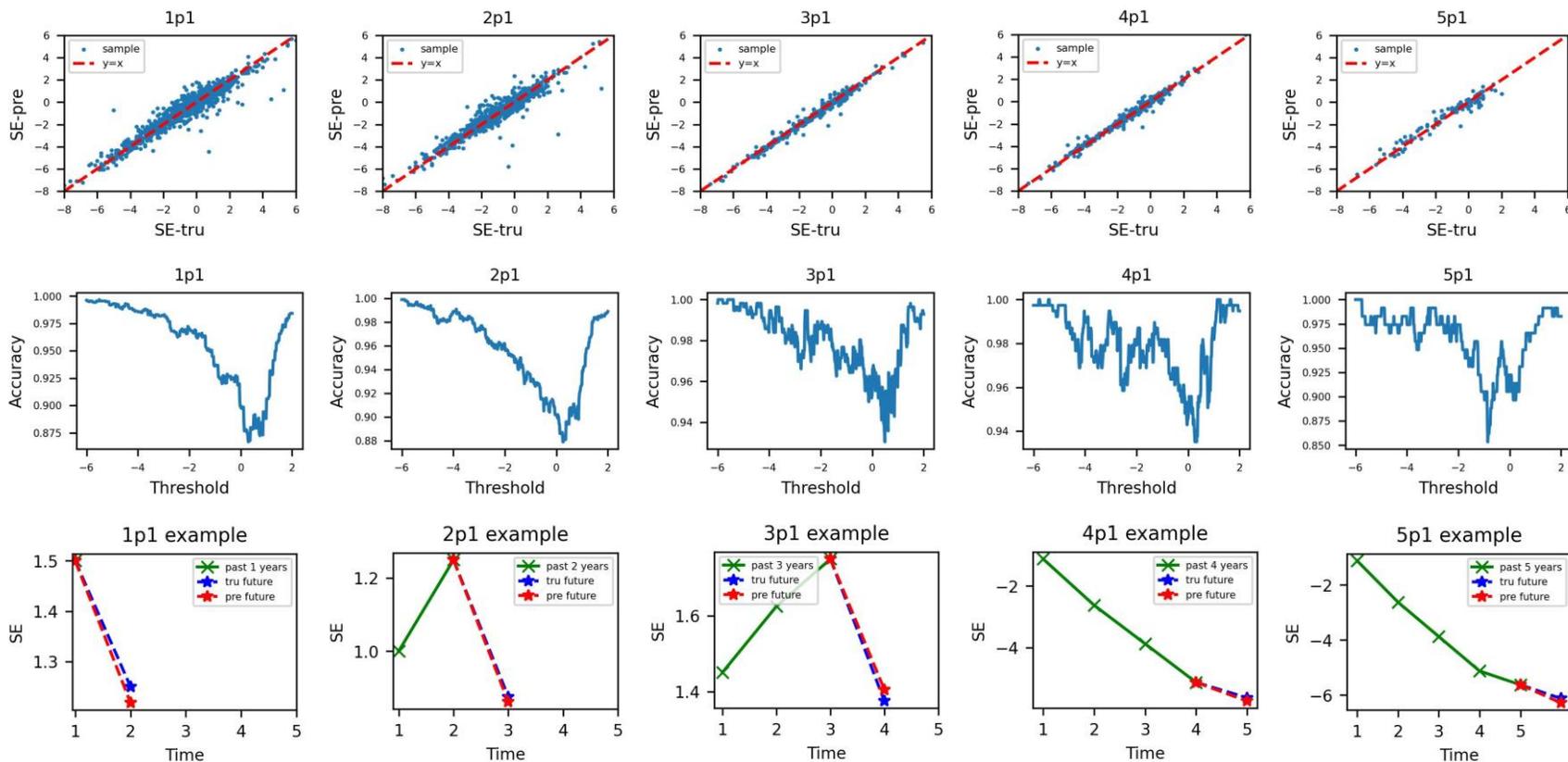

**Fig.3 Quantitative prediction of future myopia progression.** The scatterplot in the first column shows the gap between our quantitative prediction results and the true values. The curve graph in the second column categorizes the future risks using different thresholds. Lastly, the line graph in the third column presents several typical predictions of children's myopia development compared with the true results to show the prediction effect of the model on future myopia development trends. This figure provides an example of predicting one year into the future. See the supporting information for other results. The images on the right using MAE and R2 present a complete picture of differences in the model's performance in quantitatively predicting future myopia development for different lengths of input sequences and prediction sequences. (The npm at the top of the picture indicates that the n-year sequence is used to predict the future m-year sequence, and the same is true for the subsequent ones as well as in SI)

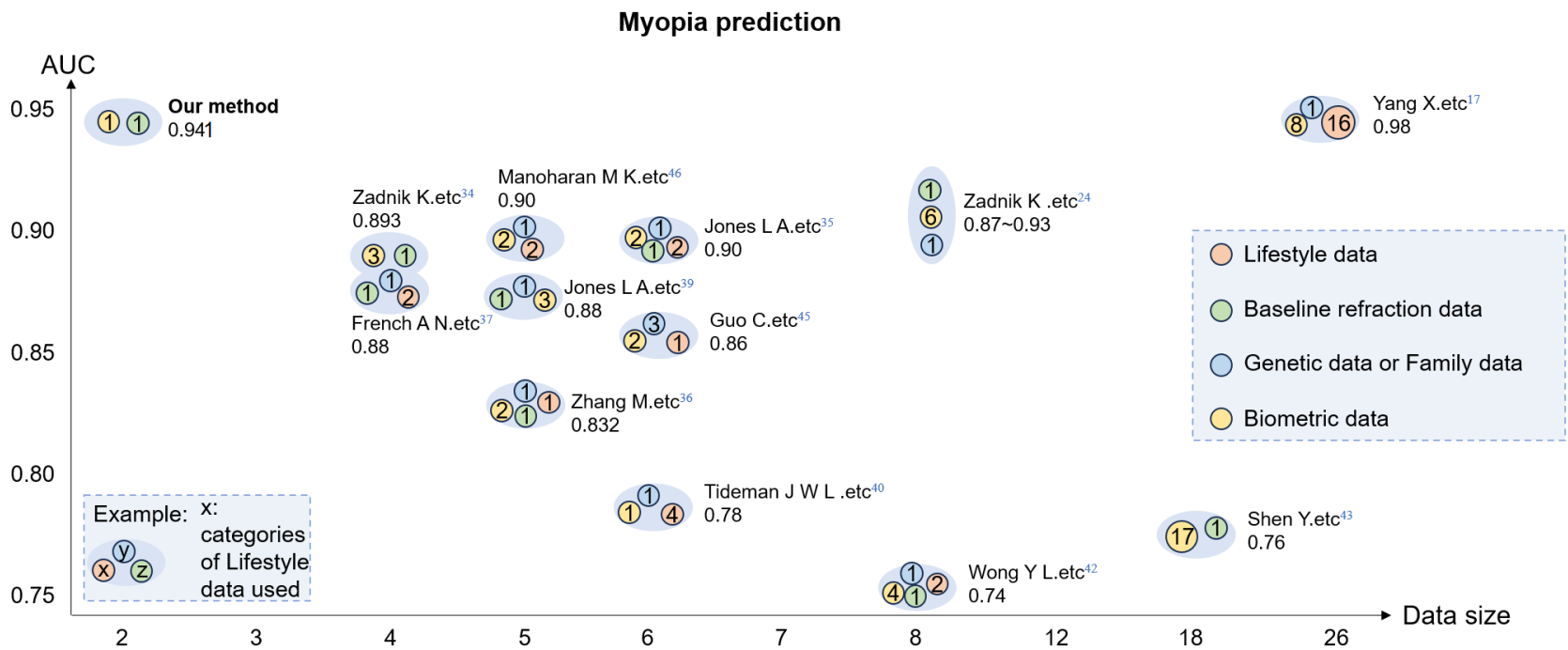

**Fig4. Comparison of method performance.** Each blue ellipse represents a study, and the different coloured circles represent the different types of data used. The numbers indicate the amount of the corresponding type of data used. The results shows that our method can achieve good performance with only a small amount of data. Performance comparison images for high myopia predictions are presented in the Supplementary Information.

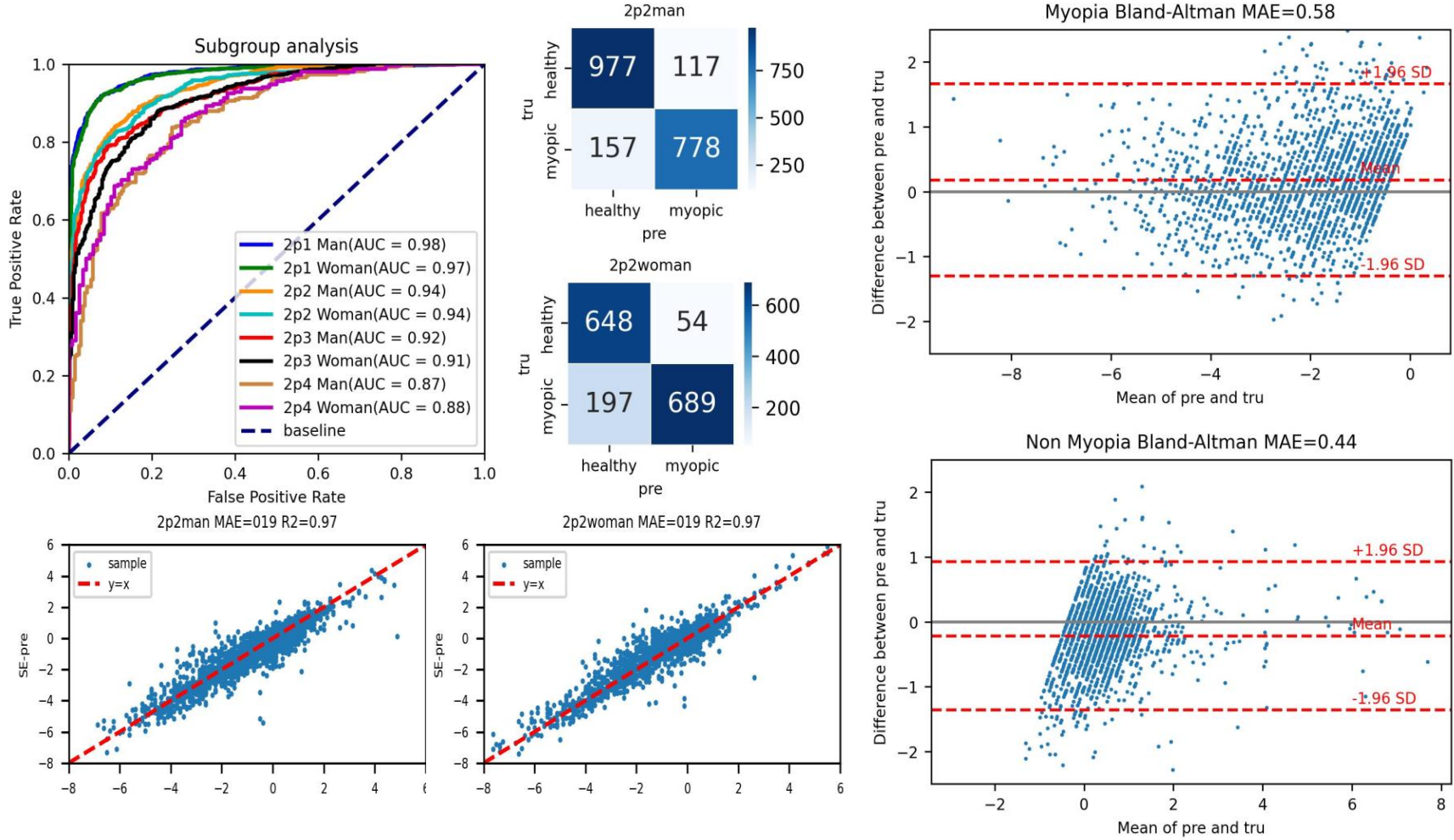

87

**Fig5. Results of subgroup analyses.** We performed a subgroup analysis of the model on male and female cohorts for a variety of
prediction sequences, and plotted the ROC curves for the two cohorts with different coloured curves. The figure presents confusion
matrices, scatter plots and MAE in the regression task. R2 metrics are shown for comparison, with two years of predictions as an
example. The two figures on the right side show the subgroup analysis of the model in the baseline myopic state, again with two-year
predictions as an example. Bland-Altman plots are also plotted as a comparison.

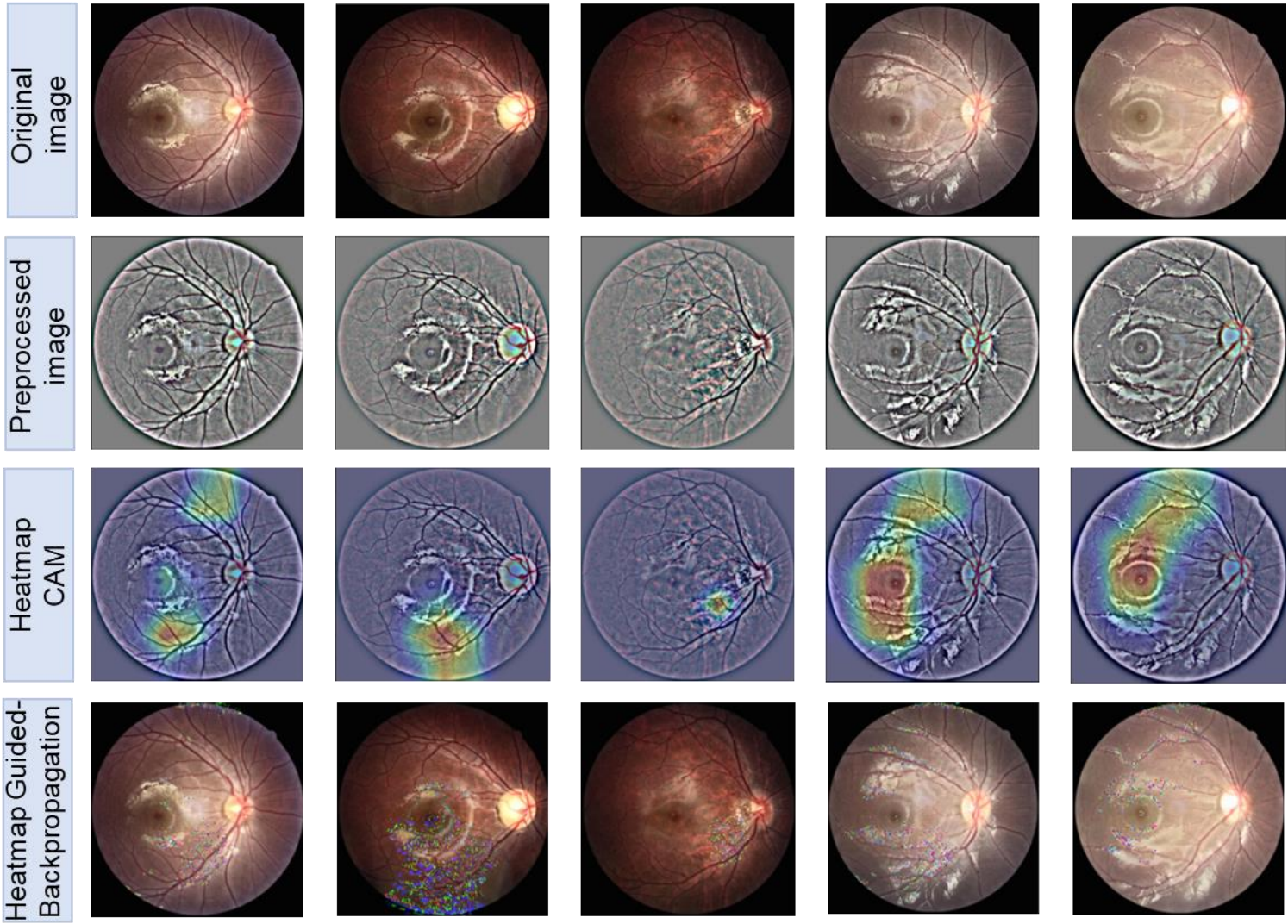

**Fig6. Examples of pre-processed images and heatmaps.** Five typical images were selected to generate the heatmap. The first three rows are examples of rapid progression (a myopia shift of more than 1.0 D/year) and the last two rows present a slower progression of myopia (a myopia shift of less than 1.0 D/year). The first column is the original image, the second column is the pre-processed image, the third column is the model's region of-interest heatmap generated using Grad CAM, and the fourth column is the fine model's region of interest heatmap generated using Guided Backpropagation.

**Supporting Information**

**Authors**

**Mengtian Kang[1,10], Yansong Hu[2,10], Shuo Gao[2,10], Ankang Zhou[2], Yuanyuan Liu[3], Hongbei Meng[2], Xuemeng Li[2], Shengbo Wang[2], Xuhang Chen[4], Hubin Zhao[5], Jing Fu[1], Guohua Hu[6], Wei Wang[7], Yanning Dai[8], Arokia Nathan[9], Peter Smielewski[4], Ningli Wang[1], Shiming Li[1]**

**Affiliations**

1 Beijing Tongren Hospital, Capital Medical University, 100005, Beijing, China

2 School of Instrumentation and Optoelectronic Engineering, Beihang University, 100191, Beijing, China

3 School of Mechanical Engineering, Beihang University, 100091, Beijing, China

4 Division of Neurosurgery, Department of Clinical Neurosciences, University of Cambridge, CB2 2PY, Cambridge, UK

5 Division of Surgery and Interventional Science, University College London, HA7 4LP, Stanmore, UK

6 Department of Electronic Engineering, The Chinese University of Hong Kong, Shatin, N. T., 999077, Hong Kong S. A. R., China

7 AI Medical Image Division, Thorough Future Inc., 100036, Beijing, China

8 AI Initiative, King Abdullah University of Science and Technology, 23955‑6900, Thuwal, Kingdom of Saudi Arabia

9 Department of Engineering, University of Cambridge, CB2 1PZ, Cambridge, UK

10 These authors contributed equally: Mengtian Kang, Yansong Hu, Shuo Gao

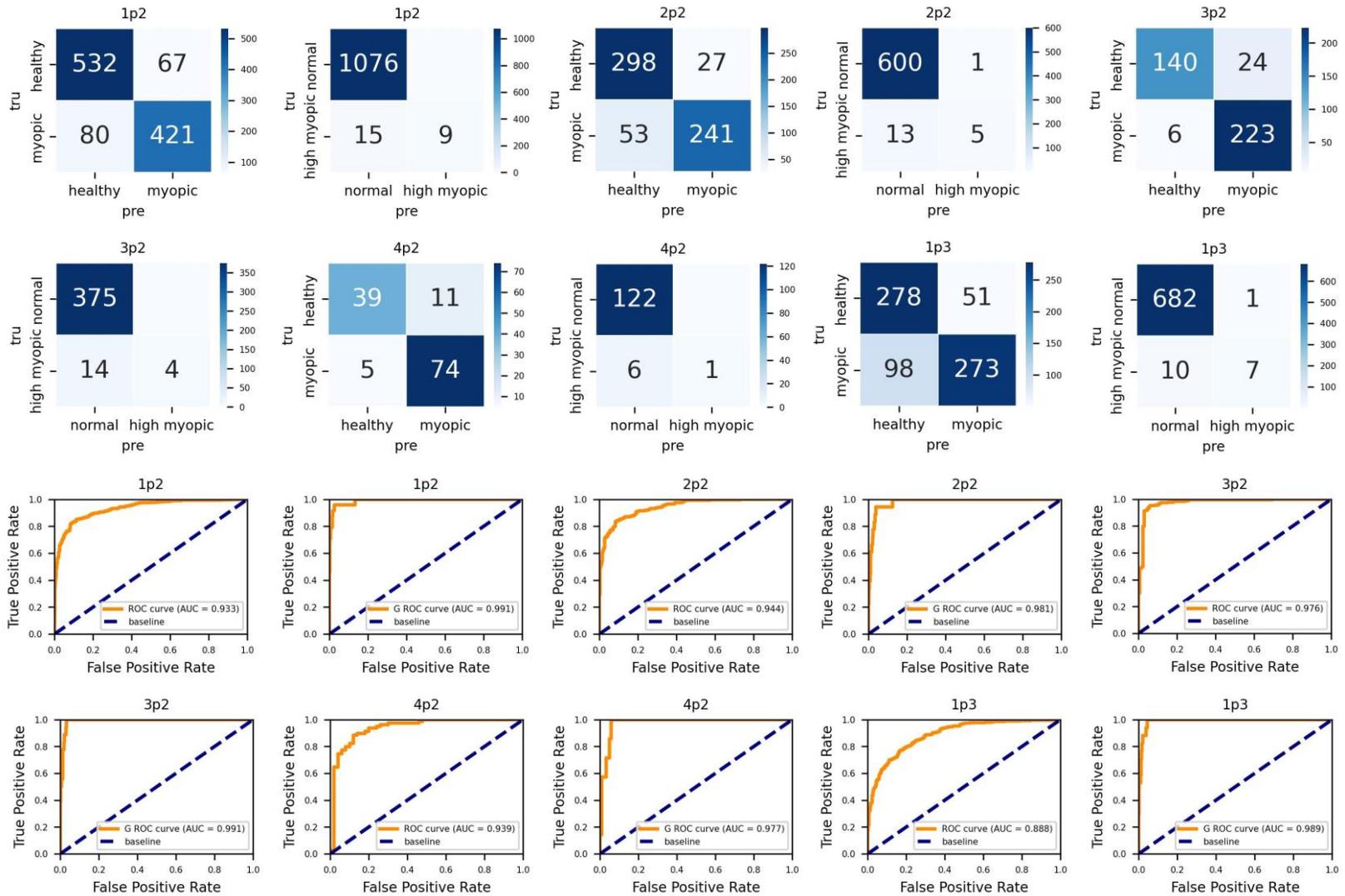

146
147 **Fig1. Supplementary Figure 1 of the results of the risk prediction for myopia and high myopia.**

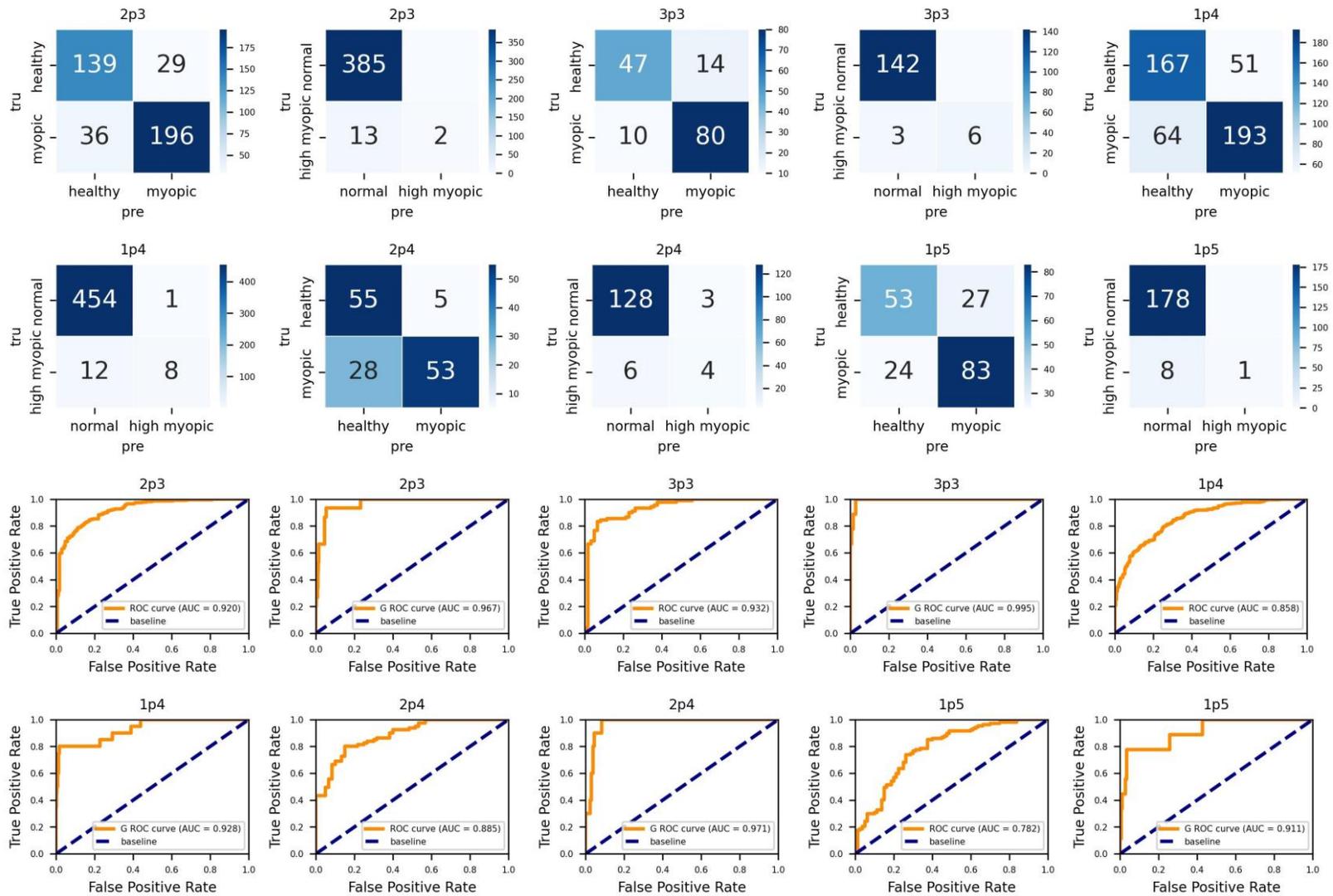

149 **Fig2. Supplementary Figure 2 of the results of the risk prediction for myopia and high myopia.**

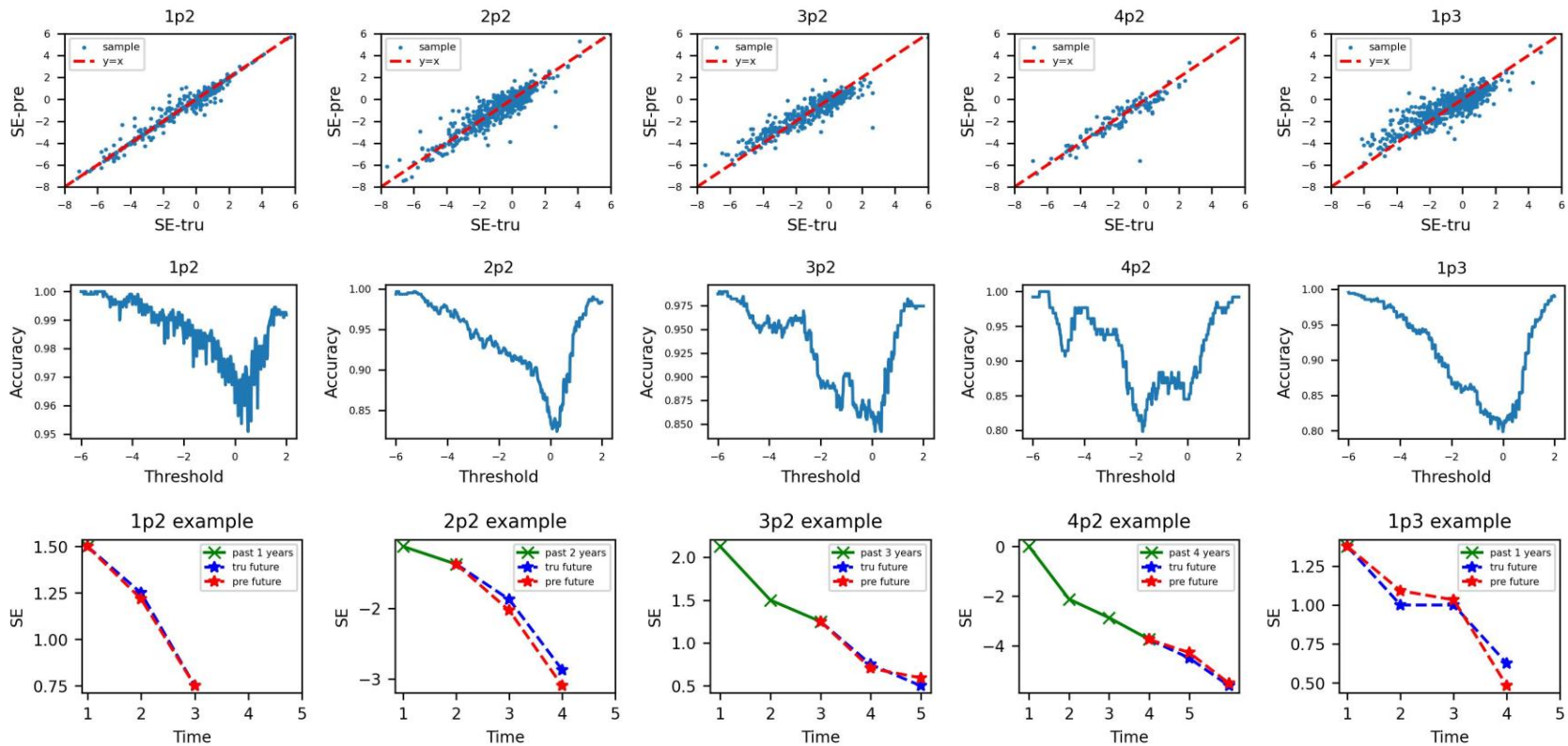

150
151 **Fig3. Supplementary Figure 1 for Quantitative Prediction of Future Myopia Development Results.**

152

153

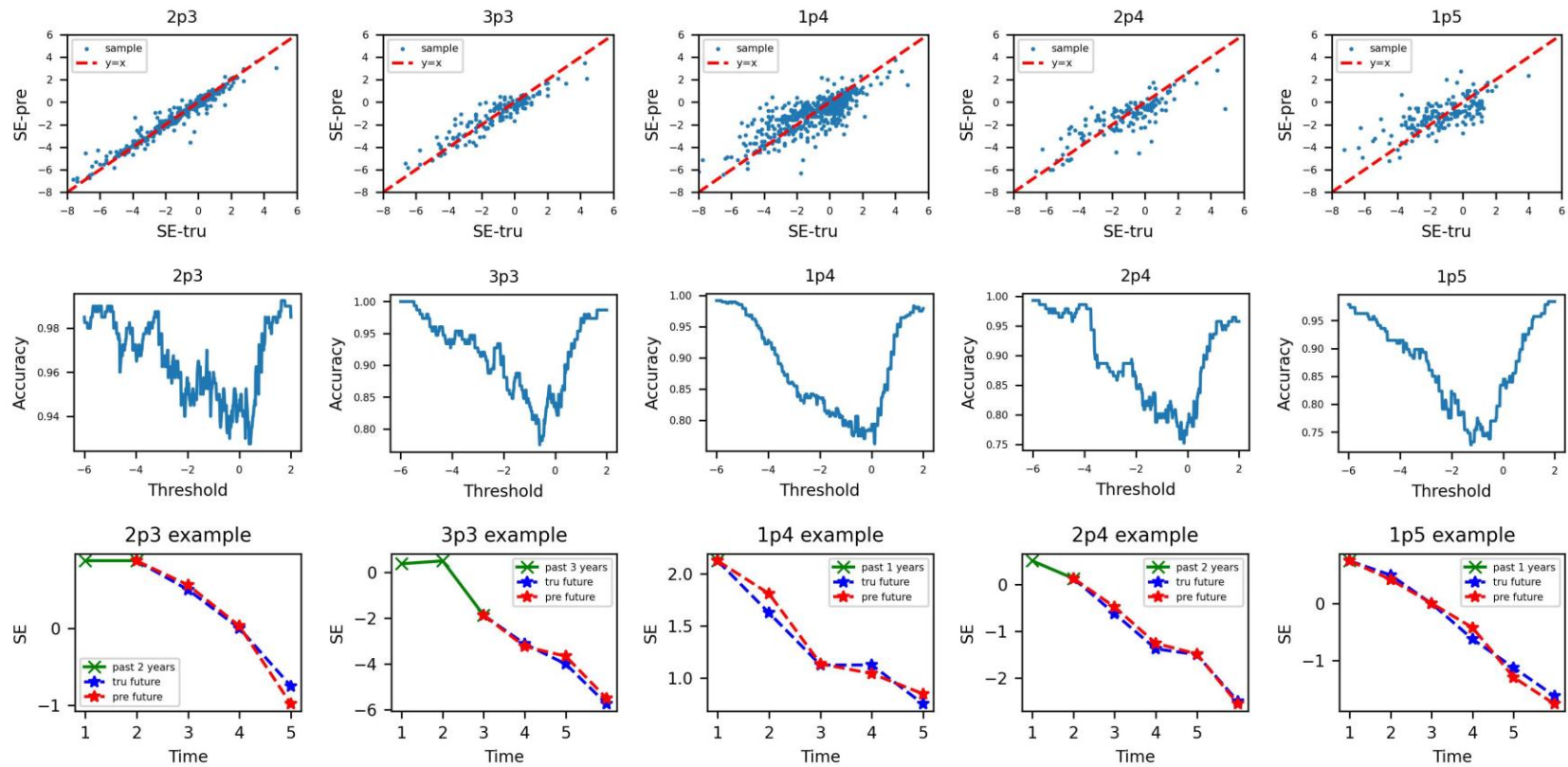

154
155  **Fig4. Supplementary Figure 2 for Quantitative Prediction of Future Myopia Development Results.**

156

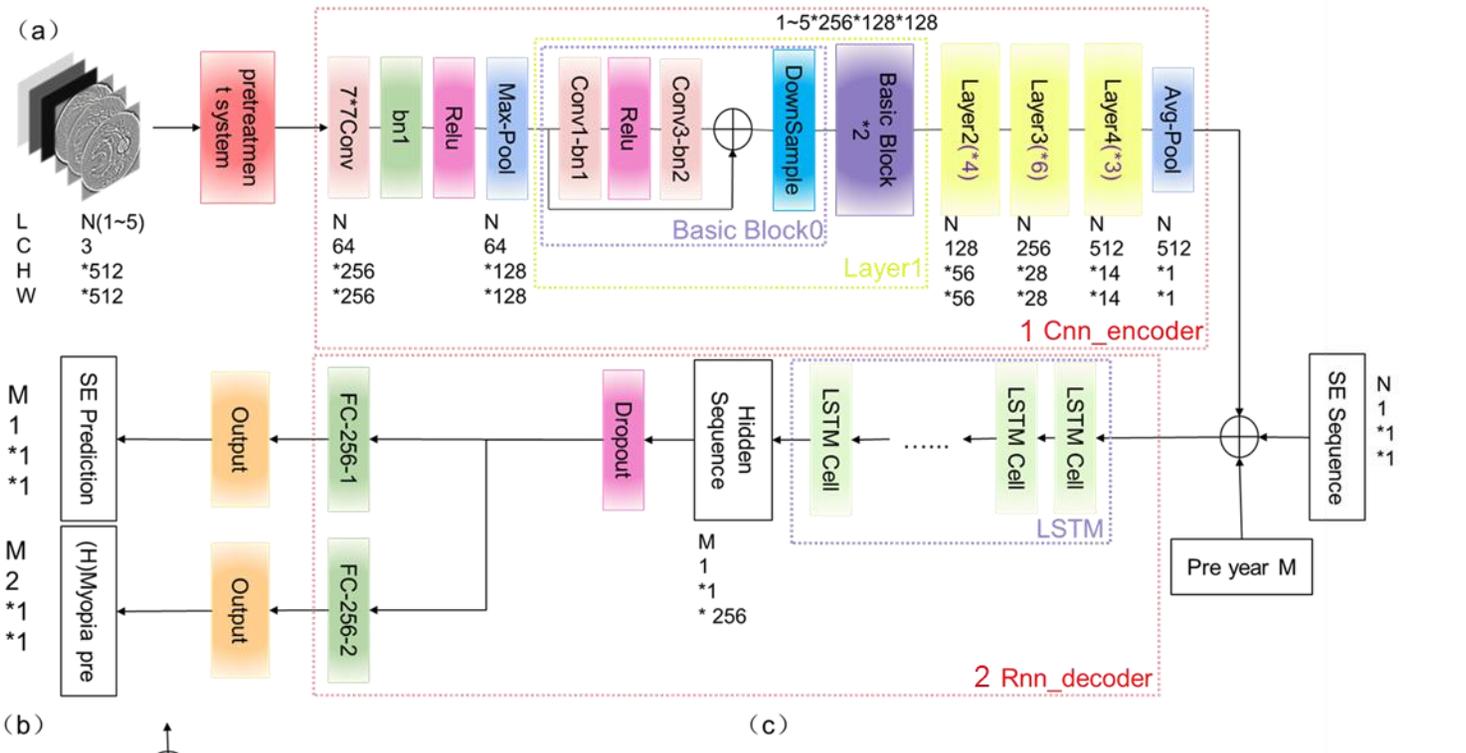

（a）

（b）

Basic Block

（c）

LSTM Cell

**Fig5. MMPN Model architecture.** The model uses the Res-Net structure in the convolutional neural network for image feature extraction and pooling to condense the features, then fuses them with the SE sequences, uses the LSTM structure in the recurrent neural network for temporal feature analysis to obtain the hidden myopia development information in the fundus image, and finally uses the fully connected layer to complete the classification or regression task.

(a)

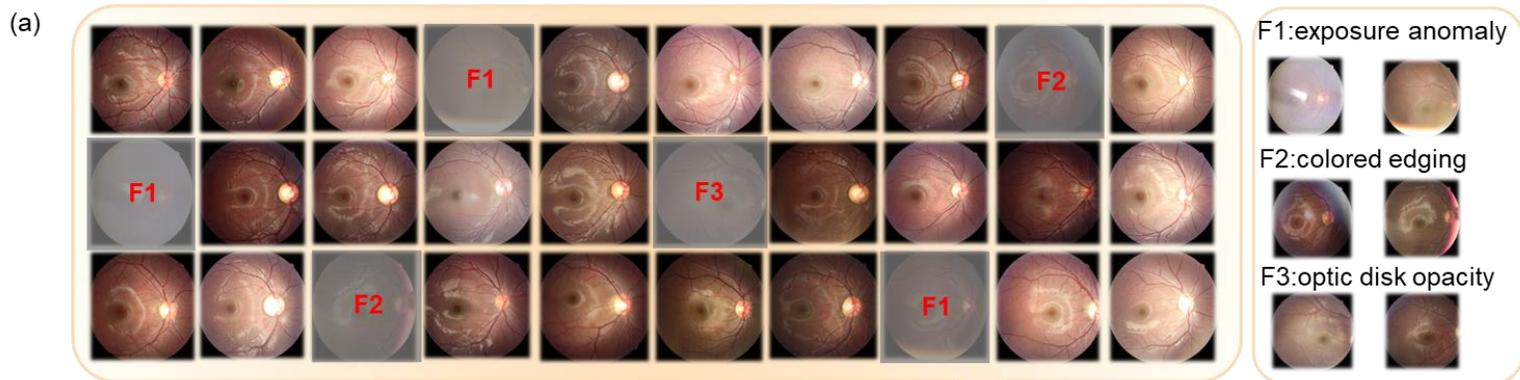

F1: exposure anomaly

F2: colored edging

F3: optic disk opacity

1600*1700*3 → **Resize** → 512*512*3 → **Screening** $HP \in [0, 2.2\%]$, $LP \in [20\%, 45\%]$, $YS \in [300, 1200]$ → 512*512*3

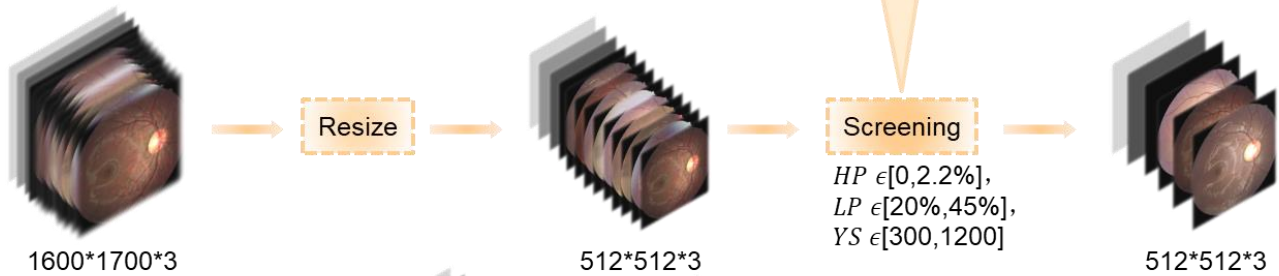

(b)

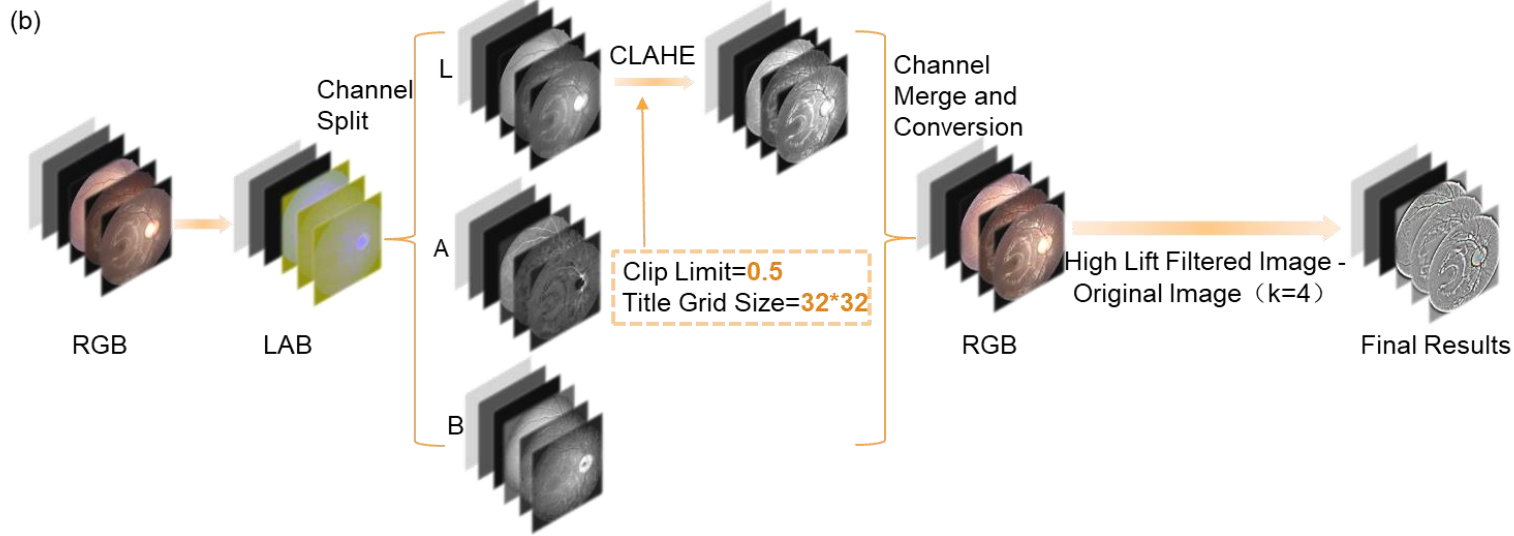

RGB → LAB → Channel Split → L, A, B → CLAHE → Channel Merge and Conversion → RGB → High Lift Filtered Image - Original Image（k=4）→ Final Results

Clip Limit=0.5
Title Grid Size=32*32

163

**Fig6. Preprocessing system.** The preprocessing system first performed basic operations such as cropping and scaling of the image, and then after manually designing three features (HP stands for the proportion of high brightness pixels, LP stands for the proportion of low brightness pixels, and YS stands for the red channel minus the grey sum of the blue channel) to complete the screening of F1, F2, F3, the three types of unqualified images, followed by the use of CLAHE enhancement of the L-channel, and then finally high lifting filtering and subtraction of the original image, and ultimately get the fundus image after the enhancement of the key physiological features. (Rotation, normalization and other operations are performed in the model training and not in the preprocessing system.)

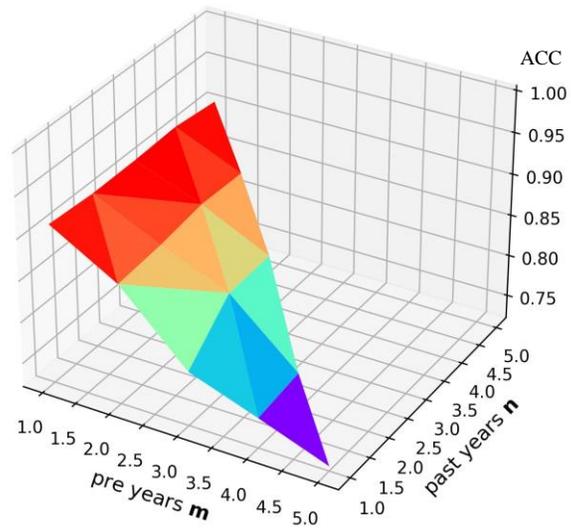

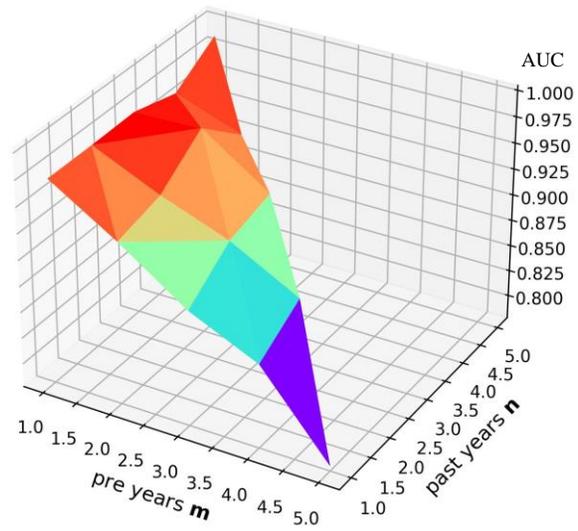

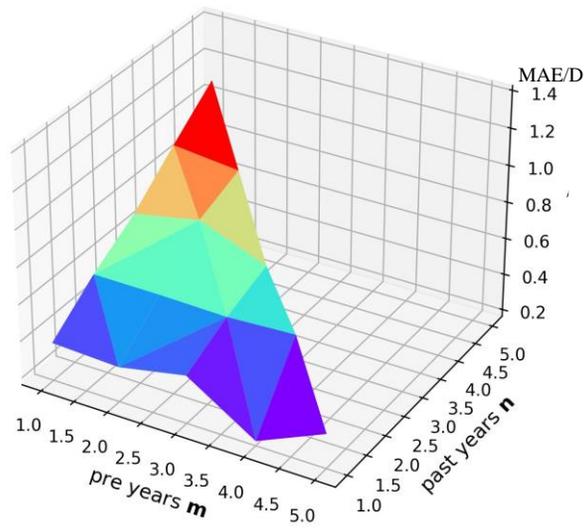

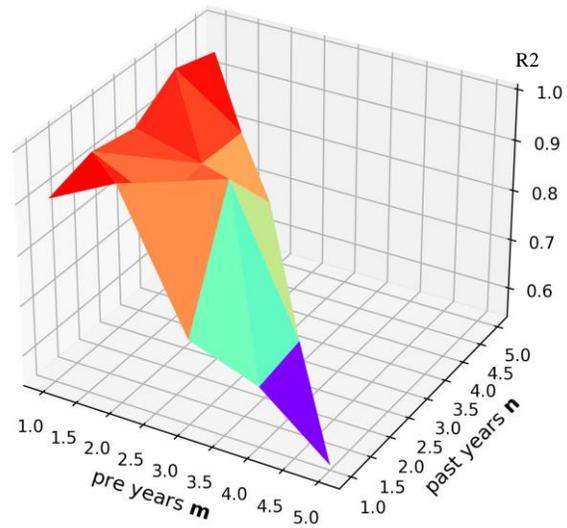

171

**Fig7. Overall performance of the nPm model.** The transformation of the performance of the nPm model for the input sequence length n versus the prediction length m is shown in the figure. In general, the performance of the prediction gets better as the input information n gets larger, and the performance of the prediction decreases as the length m to be predicted gets larger.

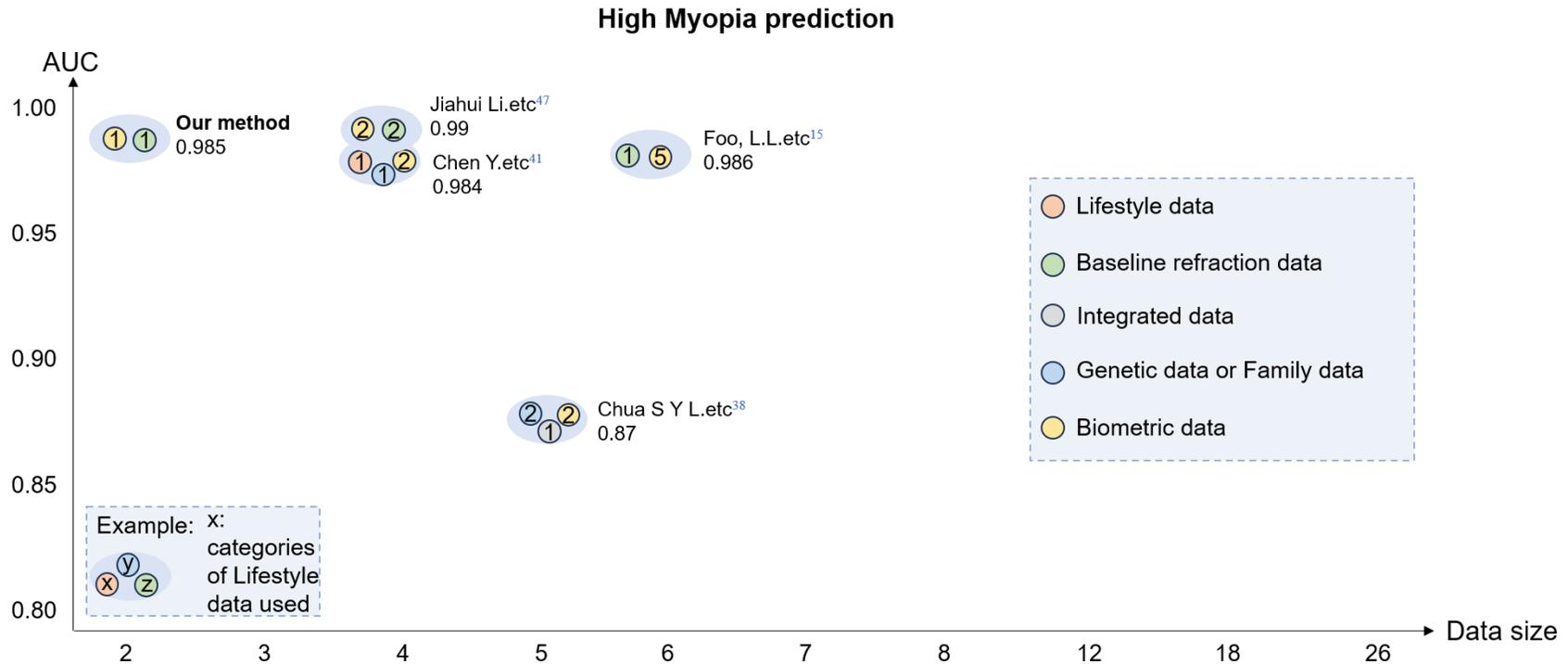

High Myopia prediction

**Fig8. Comparison of method performance.** Each blue ellipse represents a study, where the different coloured circles represent the different types of data used, and the numbers represent the amount of that type of data used. It can be seen that our method requires only a small amount of data to achieve good performance.